\documentclass[10pt,twocolumn,letterpaper,table,dvipsnames]{article}
\usepackage[pagenumbers]{iccv} % To force page numbers, e.g. for an arXiv version

\usepackage{multirow}
\usepackage{booktabs}
\usepackage{amssymb}
\usepackage{makecell}
\usepackage{float}
\usepackage{makecell} 
\usepackage{colortbl}
\definecolor{iccvblue}{rgb}{0.21,0.49,0.74}
\usepackage[pagebackref,breaklinks,colorlinks,allcolors=iccvblue]{hyperref}

\def\paperID{100} % *** Enter the Paper ID here
\def\confName{ICCV}
\def\confYear{2025}

\title{WeedSense: Multi-Task Learning for Weed Segmentation, Height Estimation, and Growth Stage Classification}

\author{
Toqi Tahamid Sarker,
Khaled R Ahmed,
Taminul Islam,
Cristiana Bernardi Rankrape,
Karla Gage\\
Southern Illinois University Carbondale, USA\\
{\tt\small \{toqitahamid.sarker, khaled.ahmed, taminul.islam, cris.rankrape, kgage\}@siu.edu}
}

\begin{document}
\maketitle
\begin{abstract}
% The ABSTRACT is to be in fully justified italicized text, at the top of the left-hand column, below the author and affiliation information.
% Use the word ``Abstract'' as the title, in 12-point Times, boldface type, centered relative to the column, initially capitalized.
% The abstract is to be in 10-point, single-spaced type.
% Leave two blank lines after the Abstract, then begin the main text.
% Look at previous \confName abstracts to get a feel for style and length.

Weed management represents a critical challenge in agriculture, significantly impacting crop yields and requiring substantial resources for control. Effective weed monitoring and analysis strategies are crucial for implementing sustainable agricultural practices and site-specific management approaches. We introduce WeedSense, a novel multi-task learning architecture for comprehensive weed analysis that jointly performs semantic segmentation, height estimation, and growth stage classification. We present a unique dataset capturing 16 weed species over an 11-week growth cycle with pixel-level annotations, height measurements, and temporal labels. WeedSense leverages a dual-path encoder incorporating Universal Inverted Bottleneck blocks and a Multi-Task Bifurcated Decoder with transformer-based feature fusion to generate multi-scale features and enable simultaneous prediction across multiple tasks. WeedSense outperforms other state-of-the-art models on our comprehensive evaluation. On our multi-task dataset, WeedSense achieves mIoU of 89.78\% for segmentation, 1.67cm MAE for height estimation, and 99.99\% accuracy for growth stage classification while maintaining real-time inference at 160 FPS. Our multitask approach achieves 3$\times$ faster inference than sequential single-task execution and uses 32.4\% fewer parameters. Please see our project page at \url{weedsense.github.io}.

\end{abstract}    
\section{Introduction}%
\label{sec:intro}

Weed management is a critical challenge in agriculture, with weeds causing global potential yield losses of 34\% across major crops~\cite{oerke2006crop}. Precise identification and monitoring of weed growth are essential for effective and sustainable agricultural practices, yet traditional methods are often labor-intensive and lack the granularity needed for site-specific weed management (SSWM)~\cite{fernandez2018current,berge2008evaluation}.

The timing of weed intervention is fundamentally linked to the Critical Period of Weed Control (CPWC) - the timeframe during which crops must remain weed-free to prevent significant yield losses~\cite{zimdahl1988concept}. Since weeds emerging with or shortly after crop emergence cause substantially greater yield losses~\cite{dew1972index, swanton2020weed}, precise early growth stage identification is critical. While traditional phenological scales like BBCH-scale~\cite{lancashire1991uniform,hess1997use} are designed for crop development, weekly monitoring intervals align better with CPWC timeframes and agricultural scheduling practices. Current weed control methods face significant limitations: manual weeding is labor-intensive and impractical for large-scale operations~\cite{forcella2000rotary,donald2007between}, while excessive herbicide use has led to environmental pollution and resistance~\cite{liu2002off,spliid1998occurrence}. Precision herbicide application, enabled by accurate weed detection, offers potential for reducing chemical usage by 54\% while maintaining efficacy~\cite{heisel1999whole}, necessitating automated identification systems for real-time, spatially-precise detection.

Deep learning has advanced computer vision tasks, achieving state-of-the-art results in image classification~\cite{simonyan2014very,he2016deep}, object detection~\cite{girshick2015fast}, and semantic segmentation~\cite{zhao2017pyramid,yu2021bisenet}. In agriculture, these techniques have been increasingly applied to crop/weed classification~\cite{kamilaris2018deep}, disease detection~\cite{mohanty2016using,goshika2023deep,ferentinos2018deep}, and yield prediction~\cite{khaki2019crop}. 
Computer vision, particularly deep learning, offers a promising avenue for automating weed monitoring, enabling rapid and accurate weed detection~\cite{haug2015crop,hasan2021survey}, growth stage classification~\cite{teimouri2018weed,almalky2023deep}, and plant height estimation~\cite{nidamanuri2024deep}.

\begin{figure*}[tpb]
    \centering
    \captionsetup{belowskip=5pt}
    \includegraphics[width=\linewidth]{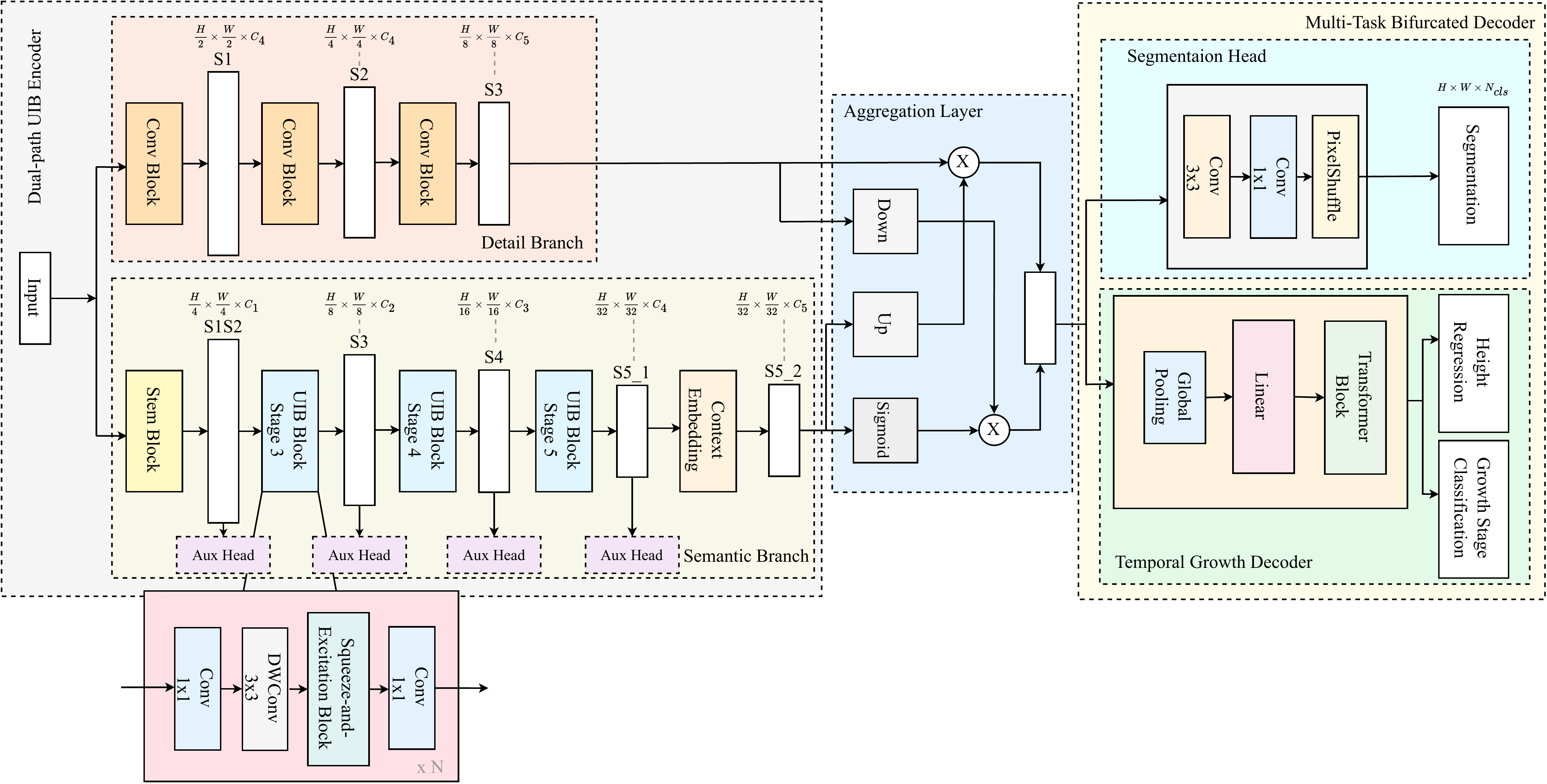}
    \caption{The proposed WeedSense framework consists of three main modules: A Dual-path UIB Encoder with parallel Detail and Semantic branches to extract multi-scale features; an Aggregation Layer that fuses these features through attention-guided operations; and a Multi-Task Bifurcated Decoder that simultaneously predicts semantic segmentation masks (17 classes including background), plant height (regression), and growth stage (11 classes). ``UIB" indicates Universal Inverted Bottleneck blocks.}
    \label{fig:architecture}
    \vspace{-15pt}
\end{figure*}

For semantic segmentation, early approaches relied on hand-crafted features~\cite{woebbecke1995shape}, while modern architectures like FCNs~\cite{long2015fully}, U-Net~\cite{ronneberger2015u}, DeepLab~\cite{chen2017deeplab}, and BiSeNetV2~\cite{yu2021bisenet} have enabled more accurate pixel-wise classification. These methods have been adapted for weed segmentation~\cite{huang2018semantic,wang2020semantic} but often lack the capability to analyze growth patterns over time, limiting their utility for comprehensive weed management systems.

Plant height estimation has evolved from Structure-from-Motion~\cite{westoby2012structure} and stereo vision~\cite{kim2021stereo} approaches that reconstruct 3D plant models~\cite{jay2015field,fujiwara2022comparison} to deep learning methods~\cite{zhao2022phenotypic} that directly learn height-related features from images. Similarly, growth stage classification has progressed from manual inspection to CNNs~\cite{rasti2021crop} and recurrent networks like LSTMs~\cite{hochreiter1997long} that can model temporal dependencies~\cite{wang2022predicting}.
However, while deep learning approaches have been applied to plant height estimation in crops, their application specifically to weed height estimation represents a novel contribution, particularly when integrated with other analysis tasks.

Multi-task learning (MTL) aims to improve performance across multiple related tasks by leveraging shared representations~\cite{ruder2017overview}. While MTL has been applied to plant phenotyping~\cite{pound2017deep}, yield prediction~\cite{sun2022simultaneous}, and disease detection~\cite{keceli2022deep}, comprehensive weed analysis integrating segmentation, height regression, and growth stage classification remains largely unexplored.

To address this gap, we introduce WeedSense, which builds upon the efficient BiSeNetV2~\cite{yu2021bisenet} framework, incorporating Universal Inverted Bottleneck (UIB) blocks~\cite{qin2024mobilenetv4} in the Dual-path UIB Encoder (DUE) encoder for enhanced feature representation. A key innovation is our Multi-Task Bifurcated Decoder (MTBD) with a Temporal Growth Decoder (TGD) component, which leverages a transformer-based feature fusion mechanism to jointly learn height regression and growth stage classification from a shared feature representation. Our growth stage classification approach predicts growth stages using weekly development intervals. The TGD uses multi-head self-attention~\cite{vaswani2017attention} to capture complex relationships between visual features and growth attributes. 
To support this research, we developed a new, richly annotated dataset capturing the growth patterns of 16 distinct weed species over an 11-week period, from sprouting to flowering stage. Collected under controlled greenhouse conditions, it includes high-resolution video sequences with per-pixel semantic segmentation masks, precise weekly height measurements, and weekly growth labels. Our main contributions are:

\begin{itemize}
    \item We introduce a new dataset of 16 weed species with annotations for semantic segmentation, height regression, and growth stage classification across their complete life cycle.
    \item We propose WeedSense, a multi-task learning architecture that jointly performs semantic segmentation, height estimation, and growth stage classification.
    \item We evaluate WeedSense's performance on our dataset and compare it with other state-of-the-art models.
\end{itemize}

The remainder of this paper is organized as follows: Section 2 introduces our dataset, Section 3 details the WeedSense architecture, Section 4 presents experimental results and ablation studies, and Section 5 concludes with a discussion of implications and future work.

\vspace{-5pt}
\section{Dataset}%
\label{sec:dataset}

% Define sizing commands for growth progression
\newcommand{\growthfigsize}{0.085}
\newcommand{\growthlabelwidth}{0.02}
\newcommand{\hspacing}{-1.5pt}

\begin{figure*}[t] 
    \centering
    
    % Week labels header
    \makebox[\growthlabelwidth\textwidth][c]{\scriptsize}
    \makebox[\growthfigsize\textwidth][c]{\scriptsize Week 1}\hspace{\hspacing}
    \makebox[\growthfigsize\textwidth][c]{\scriptsize Week 2}\hspace{\hspacing}
    \makebox[\growthfigsize\textwidth][c]{\scriptsize Week 3}\hspace{\hspacing}
    \makebox[\growthfigsize\textwidth][c]{\scriptsize Week 4}\hspace{\hspacing}
    \makebox[\growthfigsize\textwidth][c]{\scriptsize Week 5}\hspace{\hspacing}
    \makebox[\growthfigsize\textwidth][c]{\scriptsize Week 6}\hspace{\hspacing}
    \makebox[\growthfigsize\textwidth][c]{\scriptsize Week 7}\hspace{\hspacing}
    \makebox[\growthfigsize\textwidth][c]{\scriptsize Week 8}\hspace{\hspacing}
    \makebox[\growthfigsize\textwidth][c]{\scriptsize Week 9}\hspace{\hspacing}
    \makebox[\growthfigsize\textwidth][c]{\scriptsize Week 10}\hspace{\hspacing}
    \makebox[\growthfigsize\textwidth][c]{\scriptsize Week 11}
    \\[0.4pt]

    % SETFA row
    \raisebox{3ex}{\rotatebox[origin=c]{90}{\textbf{SETFA}}}\hspace{0pt}
    \includegraphics[width=\growthfigsize\textwidth]{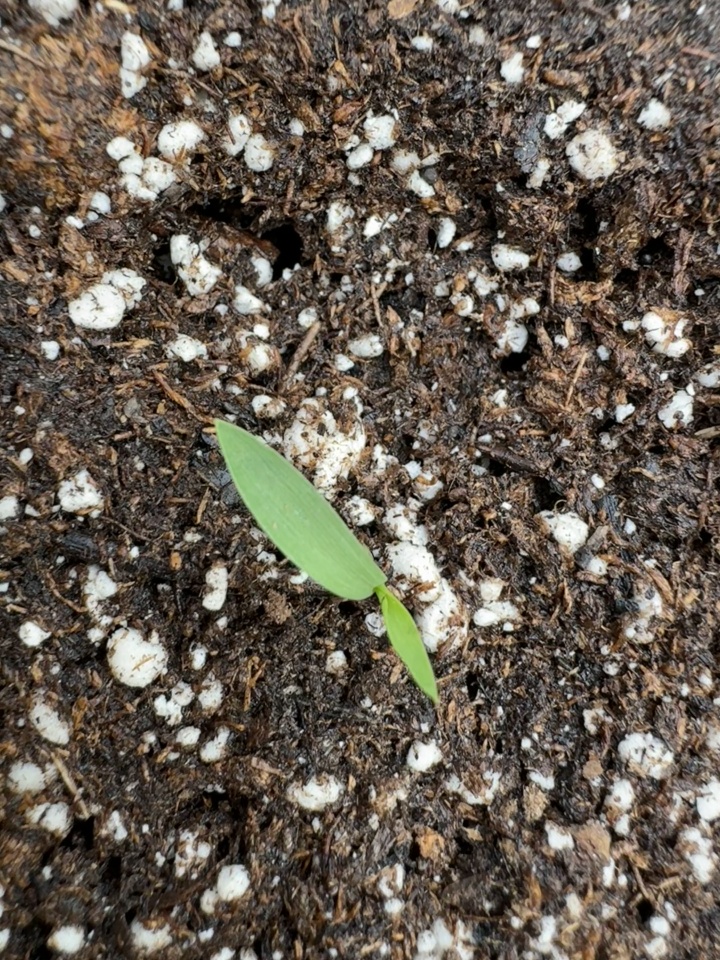}\hspace{\hspacing}
    \includegraphics[width=\growthfigsize\textwidth]{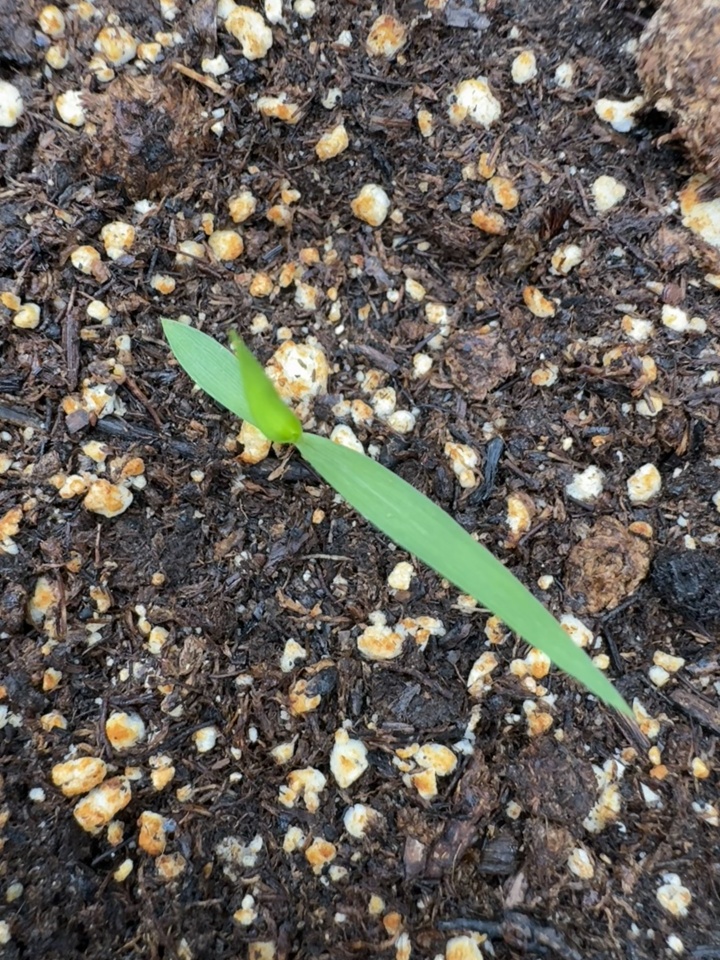}\hspace{\hspacing}
    \includegraphics[width=\growthfigsize\textwidth]{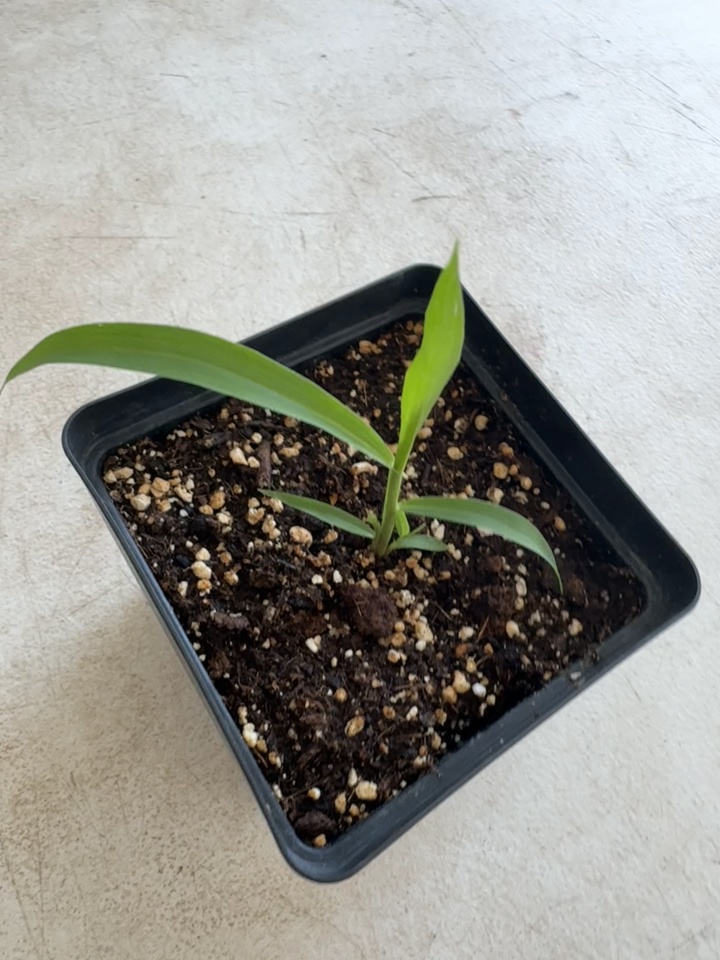}\hspace{\hspacing}
    \includegraphics[width=\growthfigsize\textwidth]{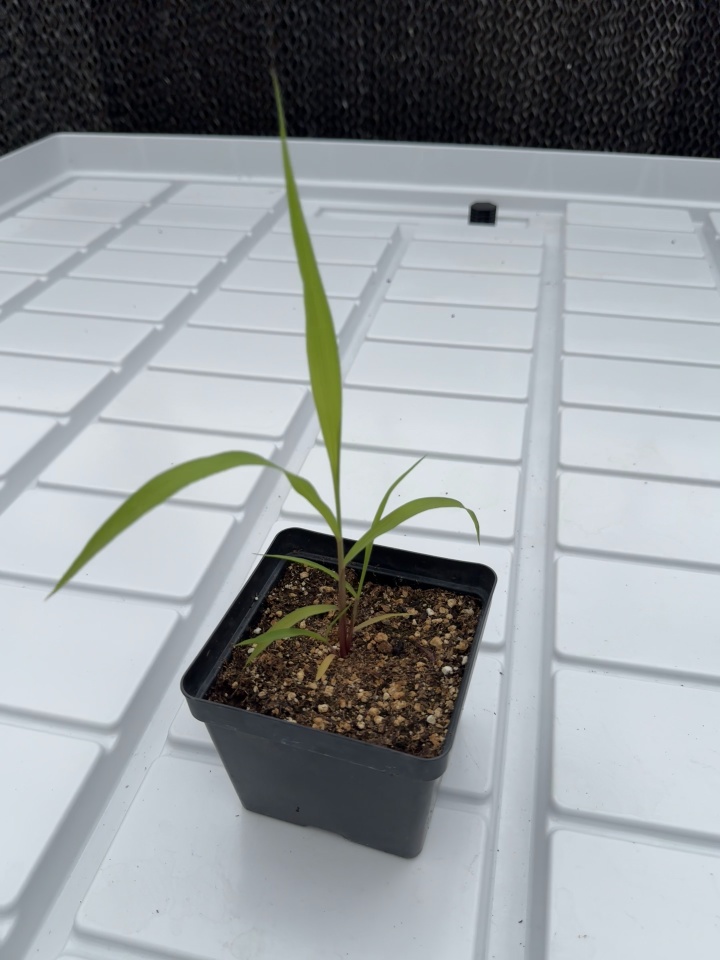}\hspace{\hspacing}
    \includegraphics[width=\growthfigsize\textwidth]{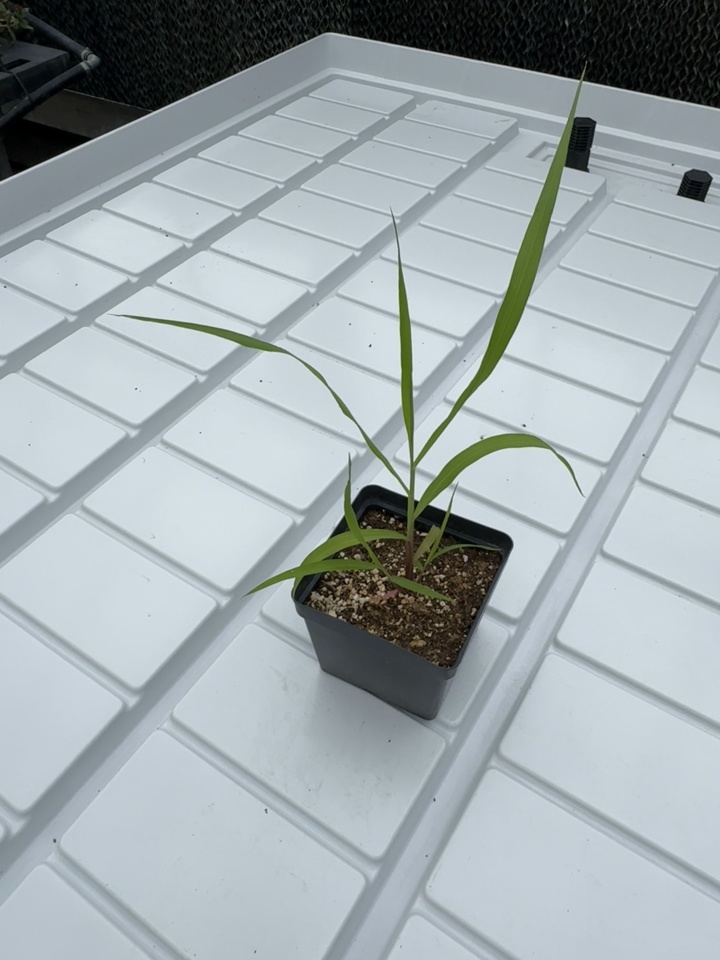}\hspace{\hspacing}
    \includegraphics[width=\growthfigsize\textwidth]{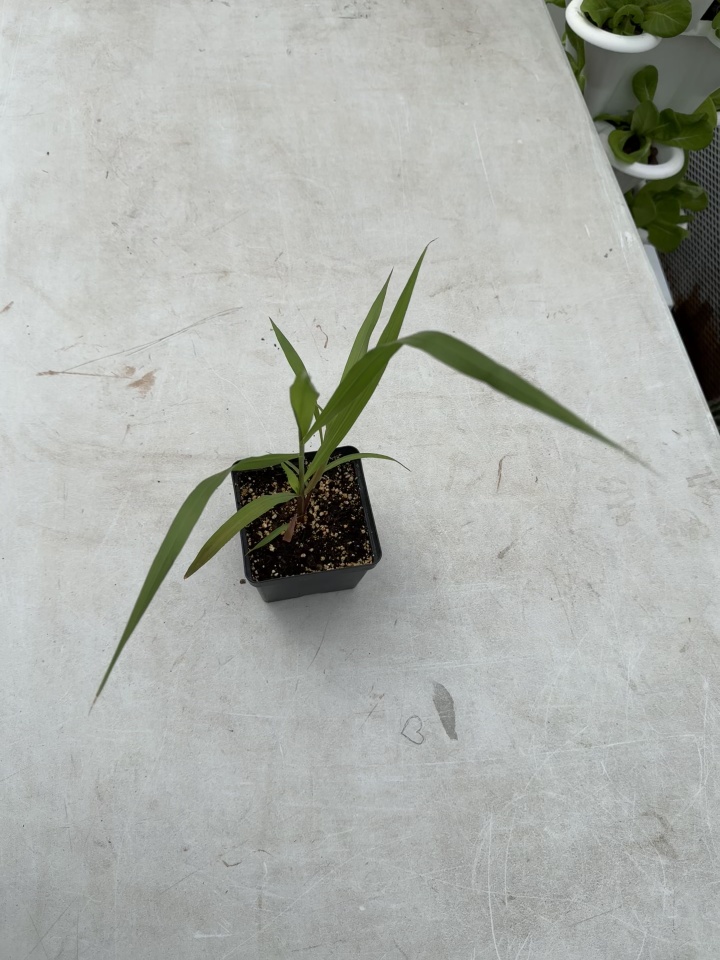}\hspace{\hspacing}
    \includegraphics[width=\growthfigsize\textwidth]{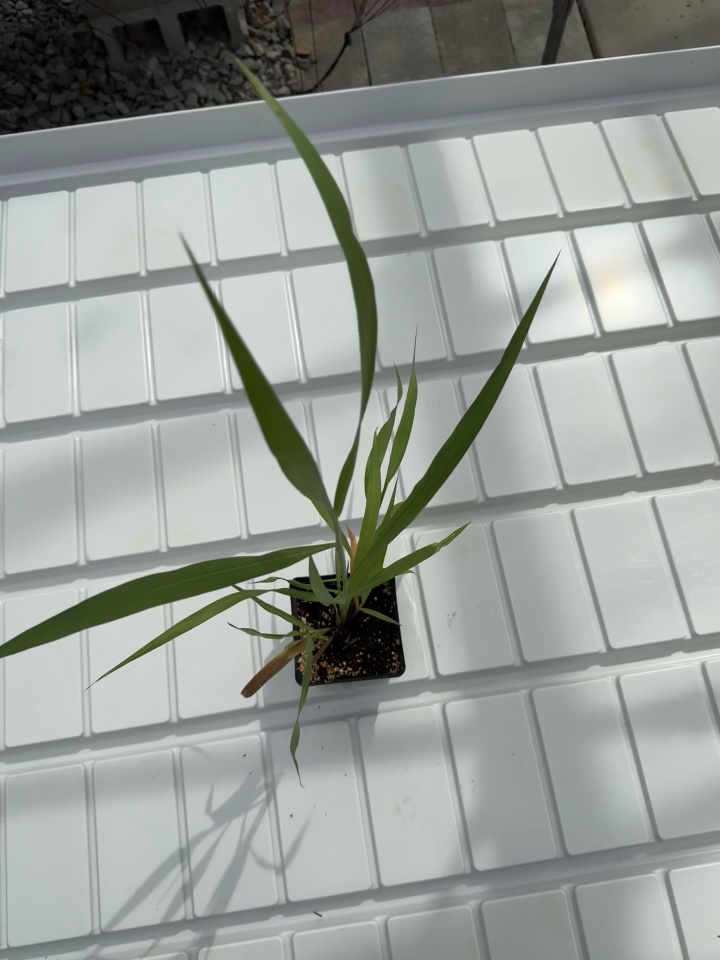}\hspace{\hspacing}
    \includegraphics[width=\growthfigsize\textwidth]{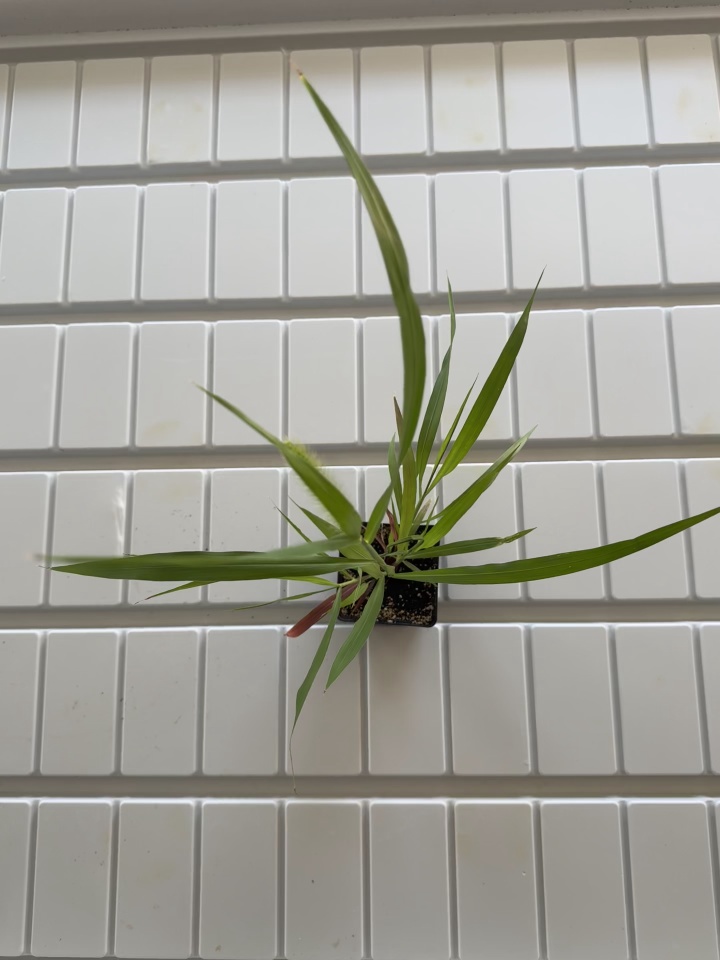}\hspace{\hspacing}
    \includegraphics[width=\growthfigsize\textwidth]{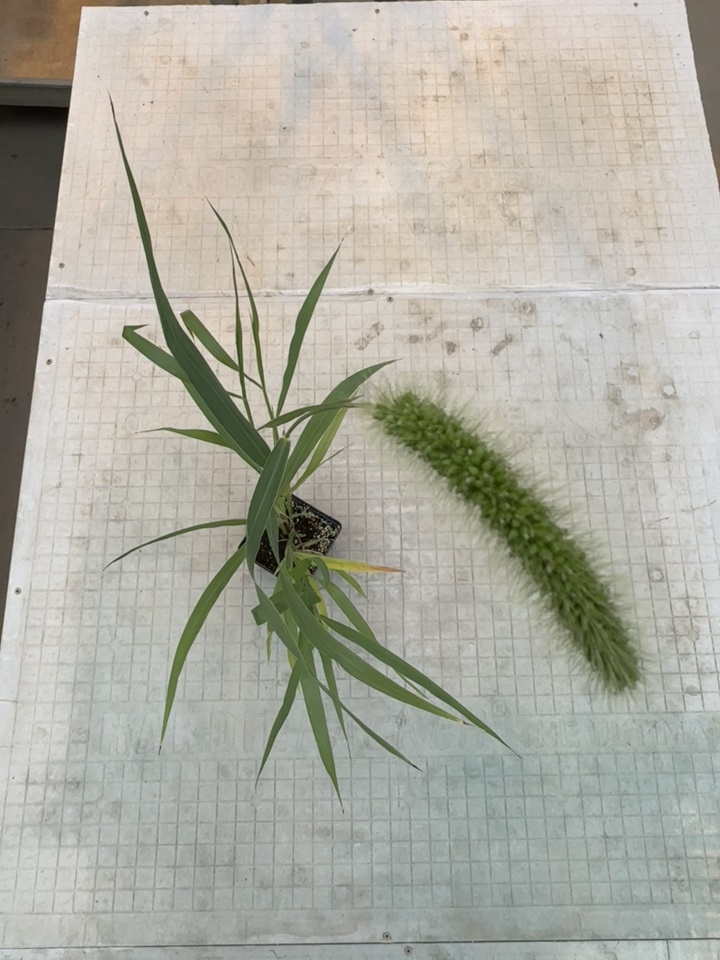}\hspace{\hspacing}
    \includegraphics[width=\growthfigsize\textwidth]{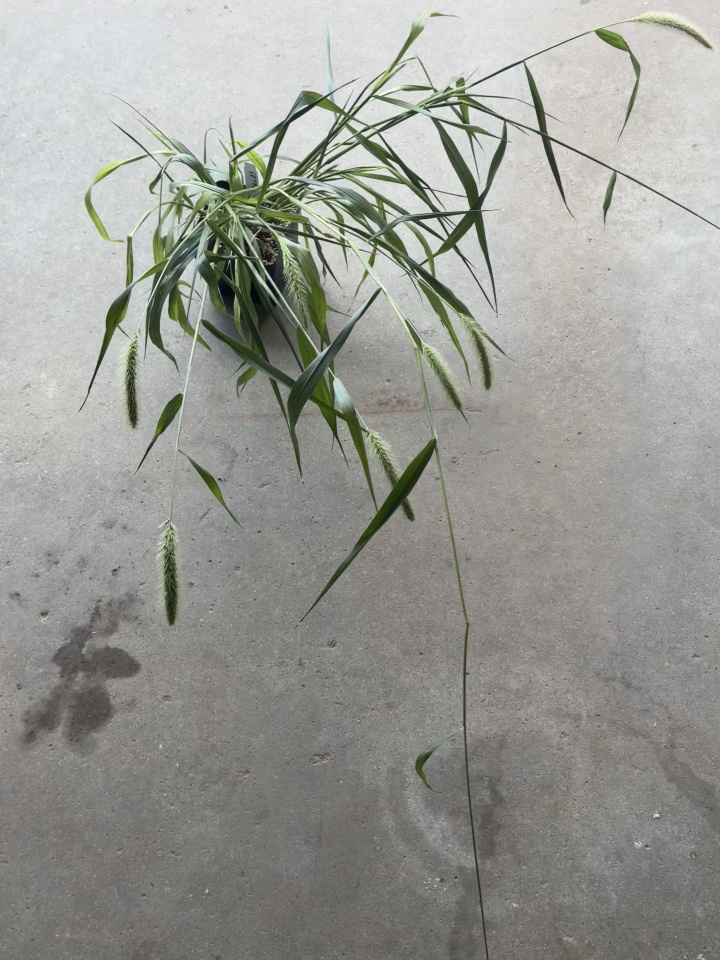}\hspace{\hspacing}
    \includegraphics[width=\growthfigsize\textwidth]{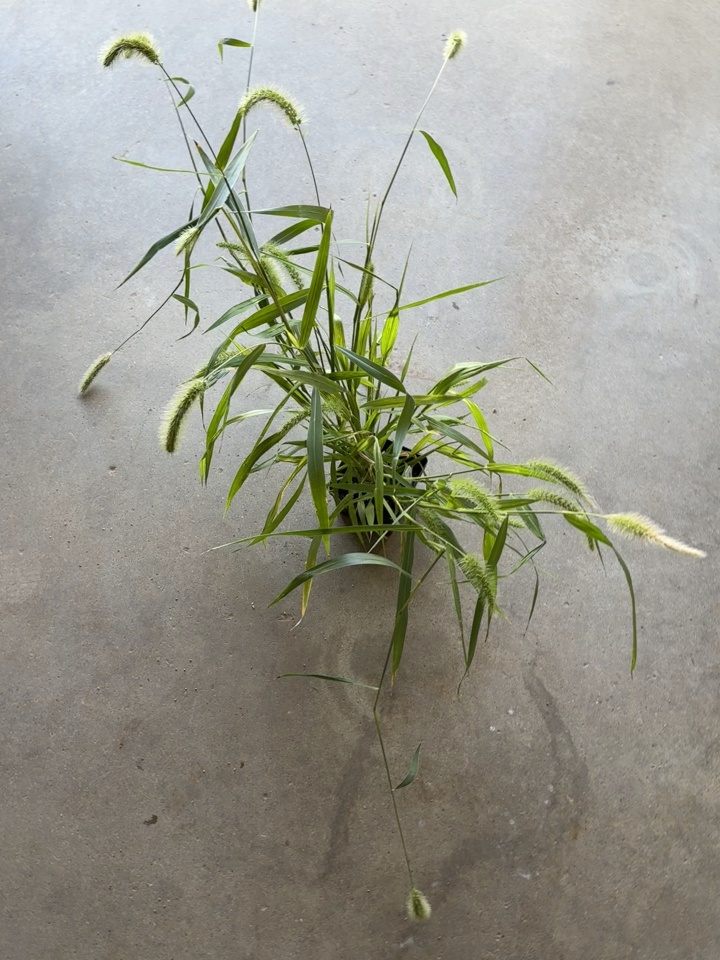}

     % AMARE row
    \raisebox{3ex}{\rotatebox[origin=c]{90}{\textbf{AMARE}}}\hspace{0pt}
    \includegraphics[width=\growthfigsize\textwidth]{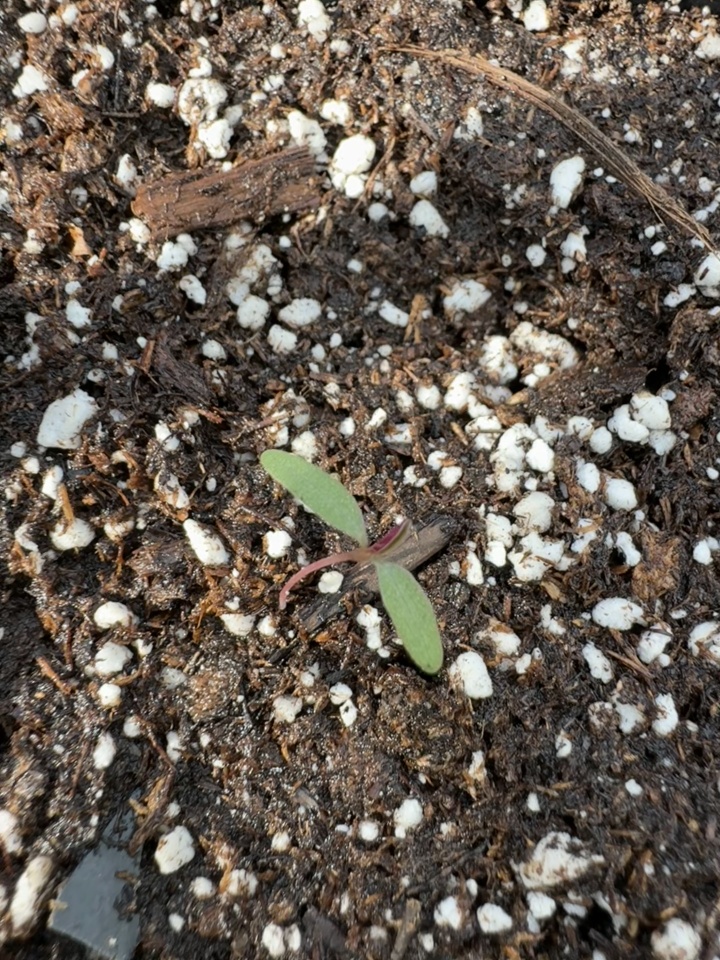}\hspace{\hspacing}
    \includegraphics[width=\growthfigsize\textwidth]{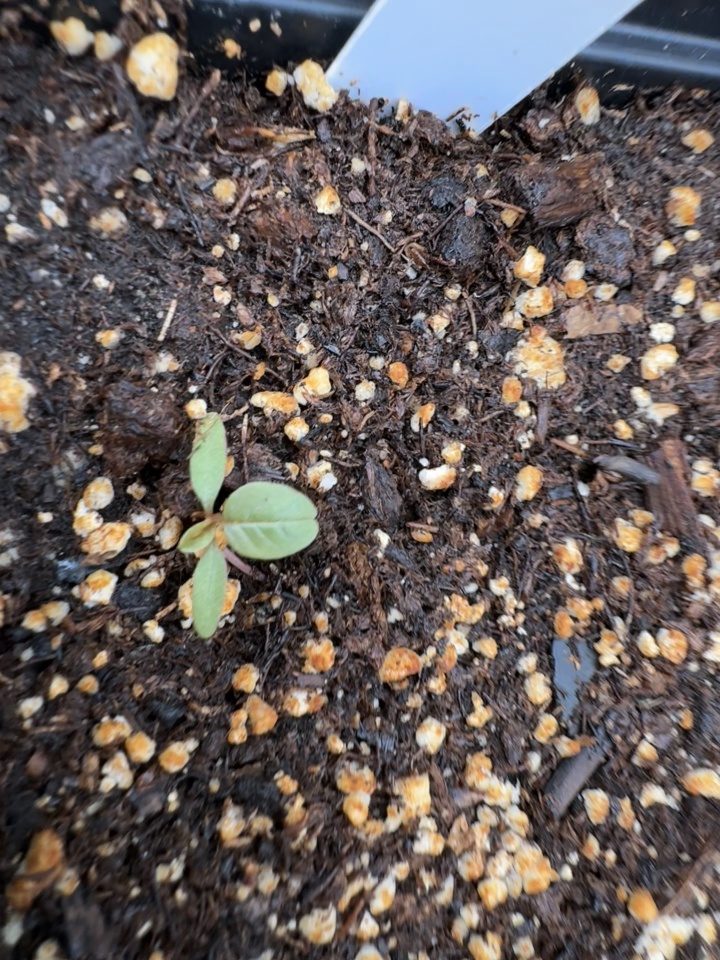}\hspace{\hspacing}
    \includegraphics[width=\growthfigsize\textwidth]{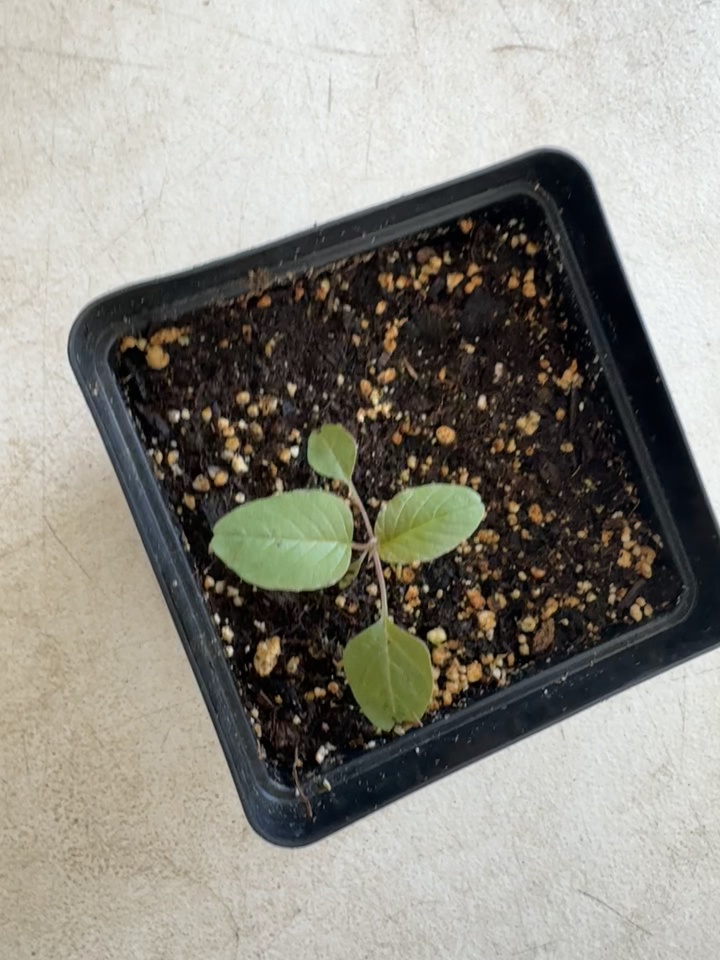}\hspace{\hspacing}
    \includegraphics[width=\growthfigsize\textwidth]{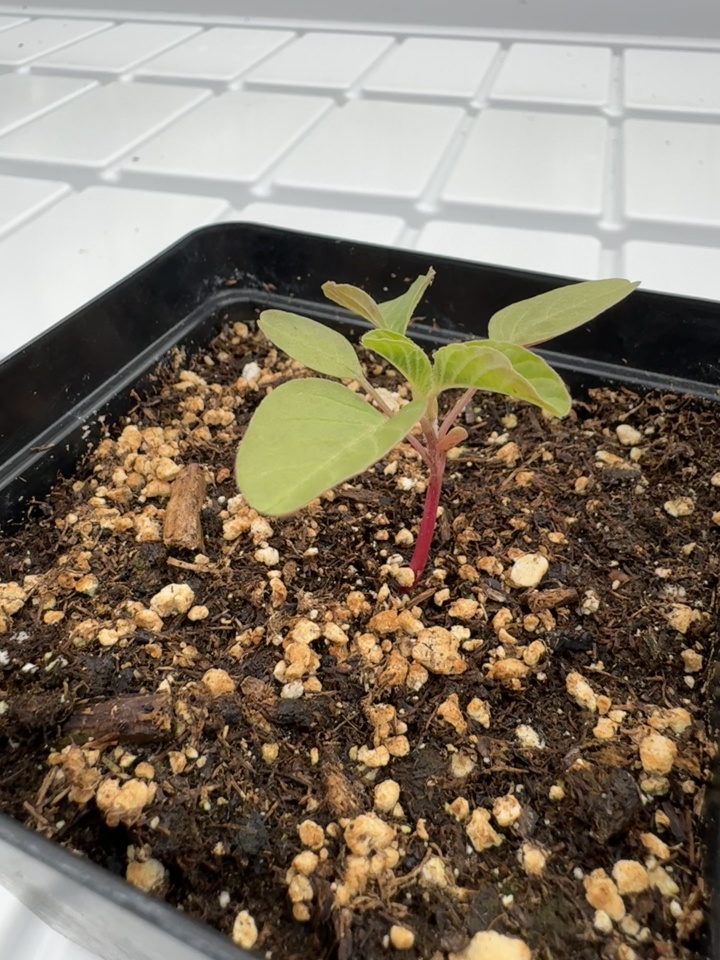}\hspace{\hspacing}
    \includegraphics[width=\growthfigsize\textwidth]{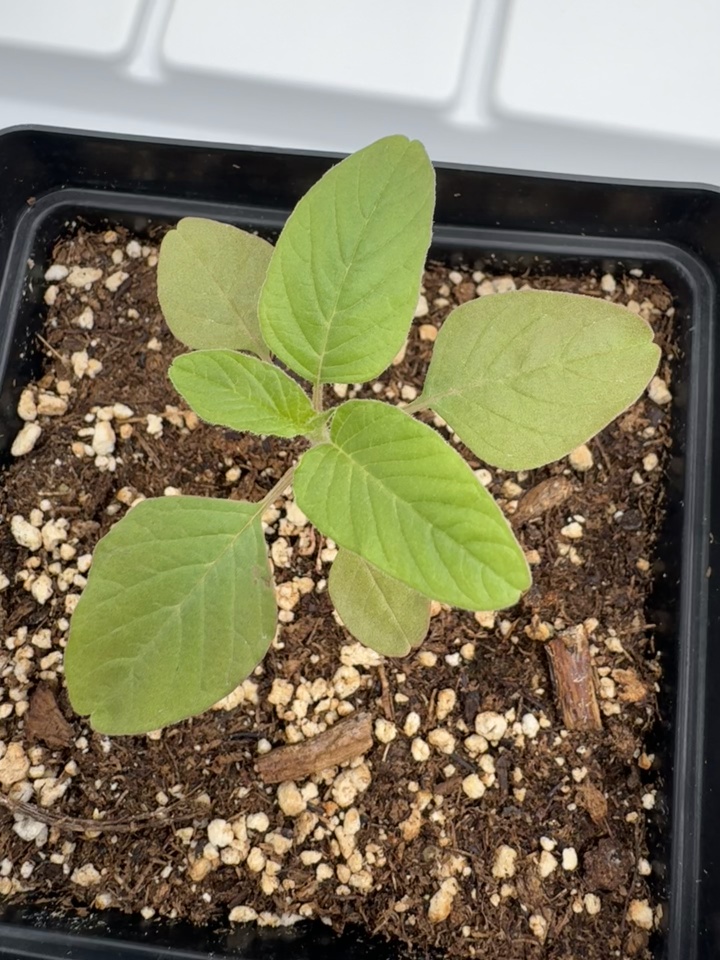}\hspace{\hspacing}
    \includegraphics[width=\growthfigsize\textwidth]{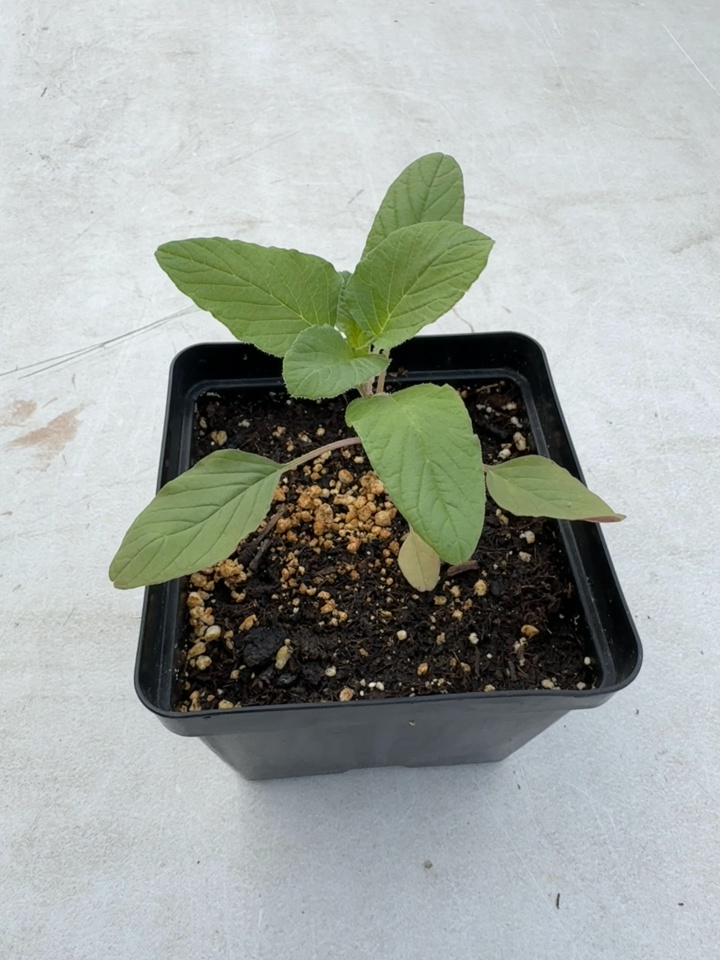}\hspace{\hspacing}
    \includegraphics[width=\growthfigsize\textwidth]{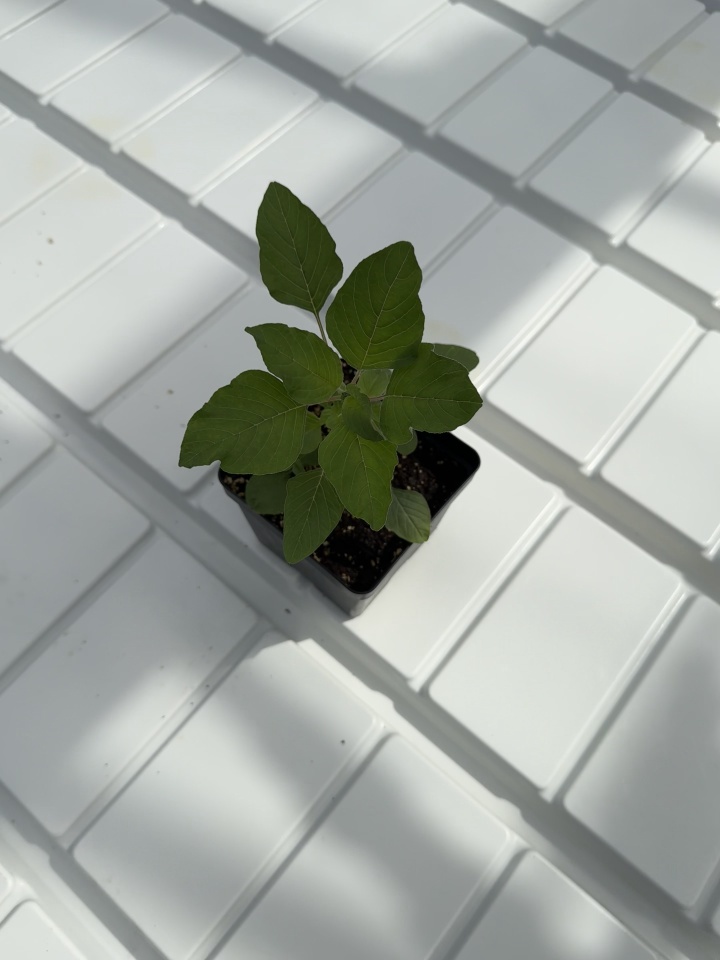}\hspace{\hspacing}
    \includegraphics[width=\growthfigsize\textwidth]{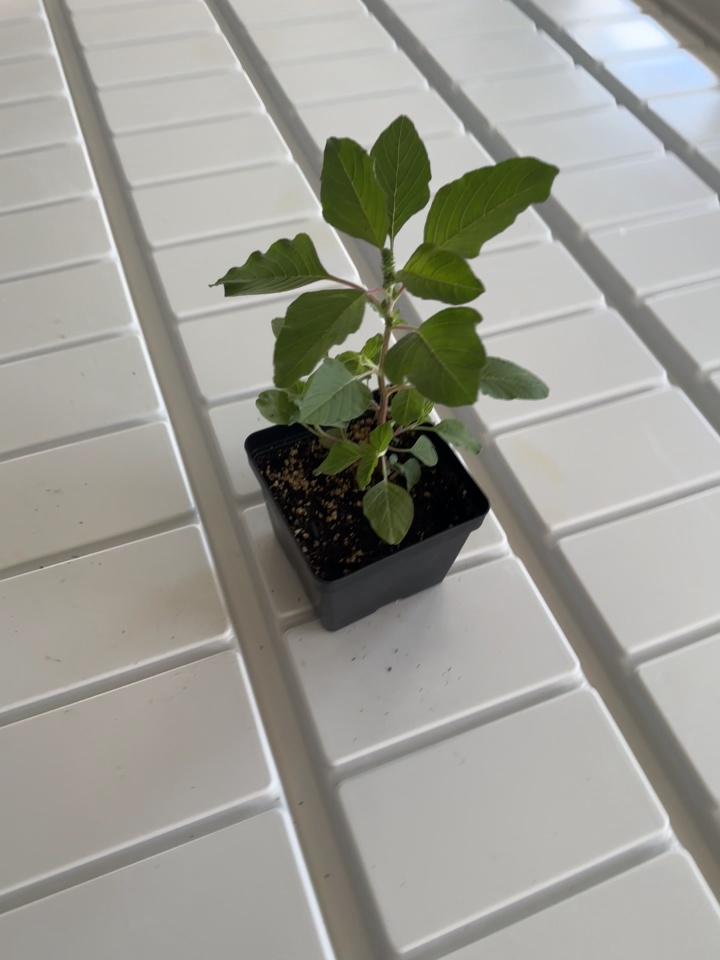}\hspace{\hspacing}
    \includegraphics[width=\growthfigsize\textwidth]{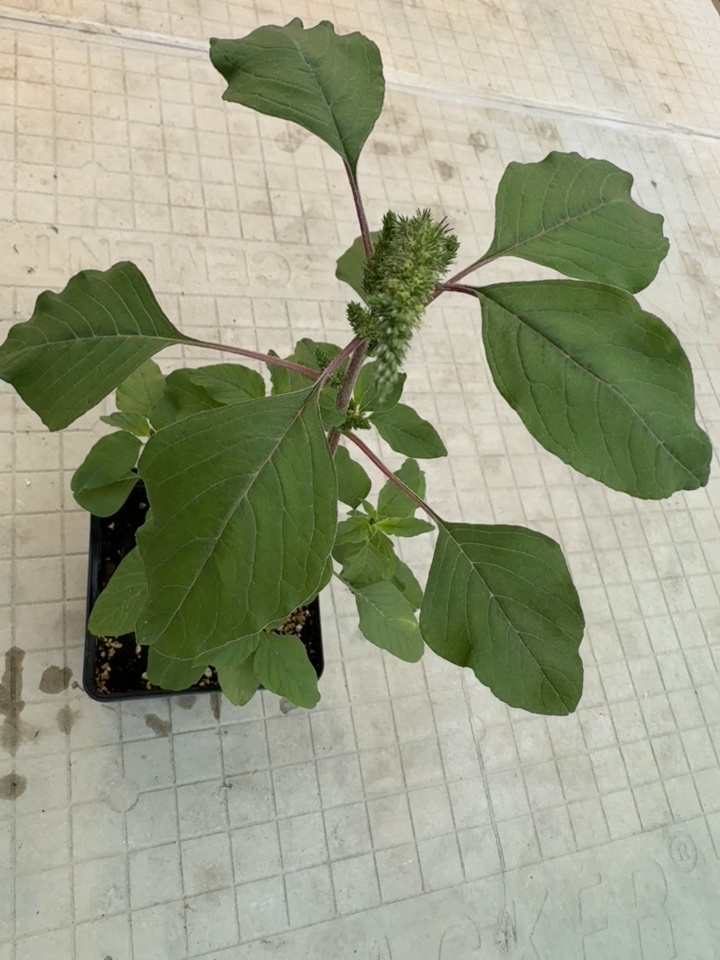}\hspace{\hspacing}
    \includegraphics[width=\growthfigsize\textwidth]{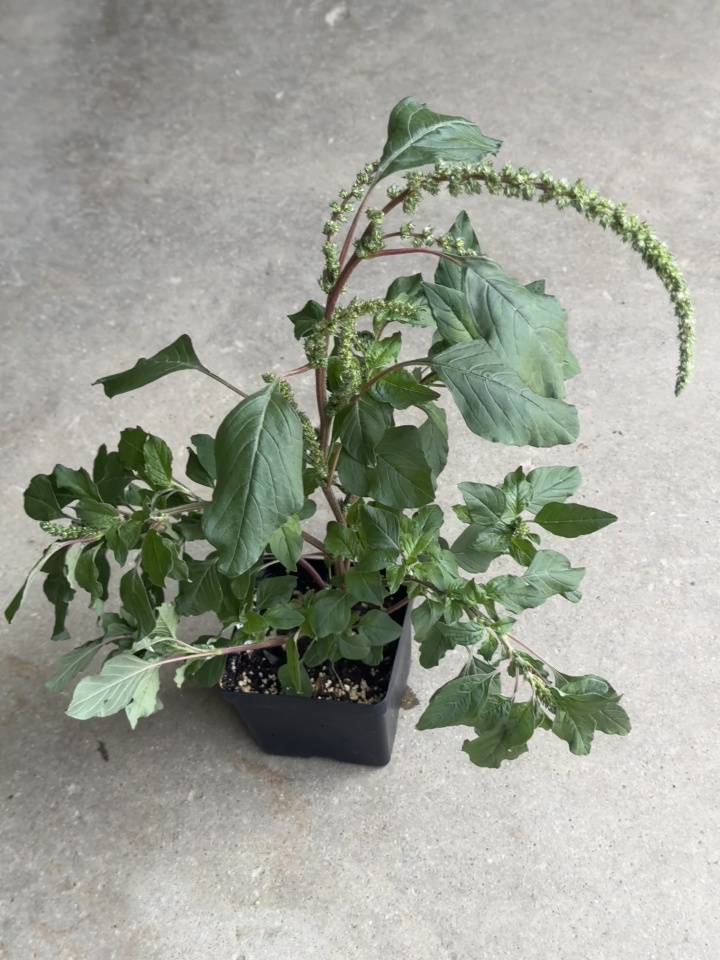}\hspace{\hspacing}
    \includegraphics[width=\growthfigsize\textwidth]{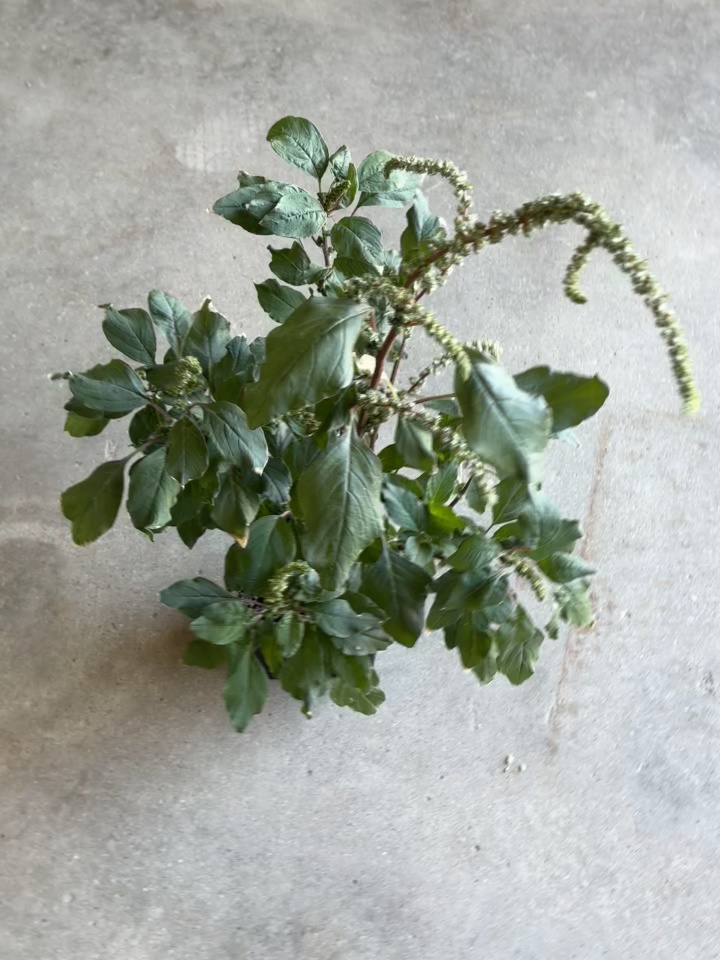}
    \\[-2pt]
    
    % ERICA row
    \raisebox{3ex}{\rotatebox[origin=c]{90}{\textbf{ERICA}}}\hspace{0pt}
    \includegraphics[width=\growthfigsize\textwidth]{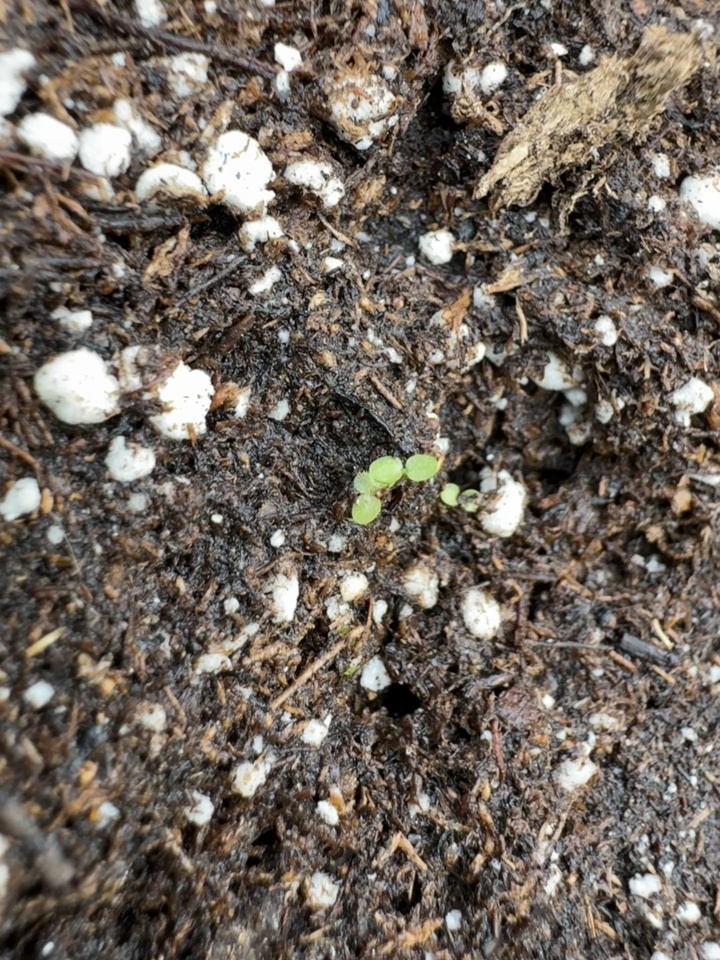}\hspace{\hspacing}
    \includegraphics[width=\growthfigsize\textwidth]{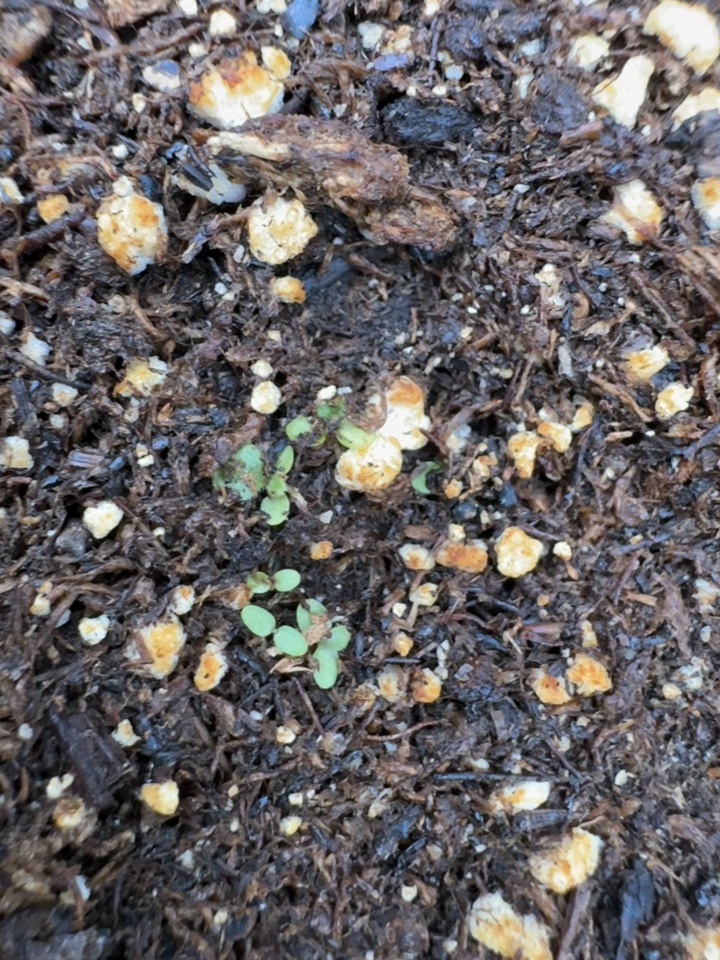}\hspace{\hspacing}
    \includegraphics[width=\growthfigsize\textwidth]{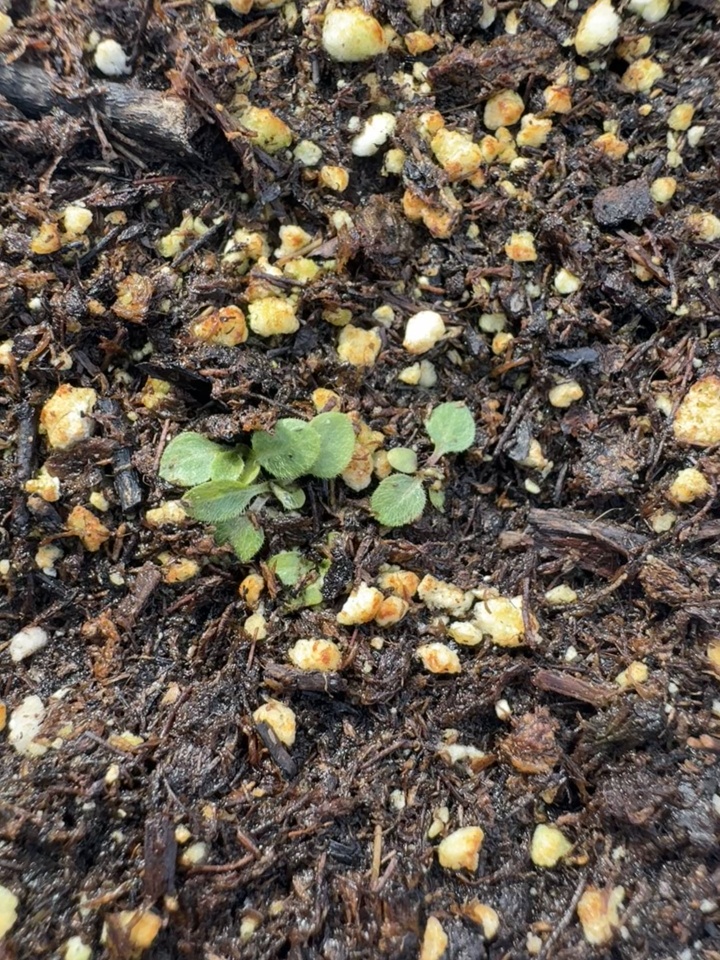}\hspace{\hspacing}
    \includegraphics[width=\growthfigsize\textwidth]{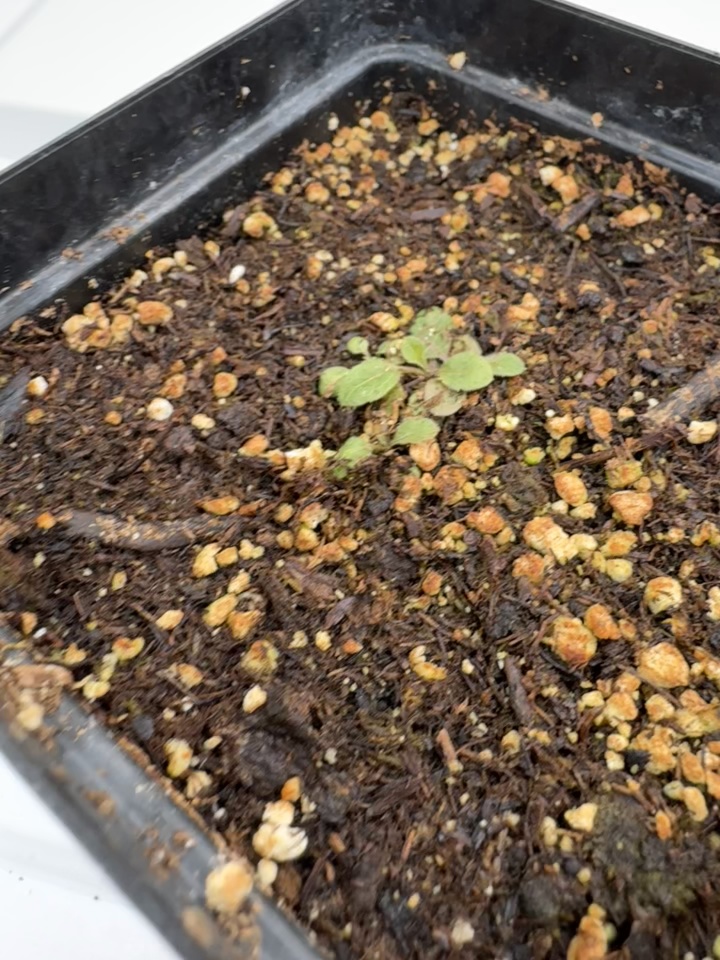}\hspace{\hspacing}
    \includegraphics[width=\growthfigsize\textwidth]{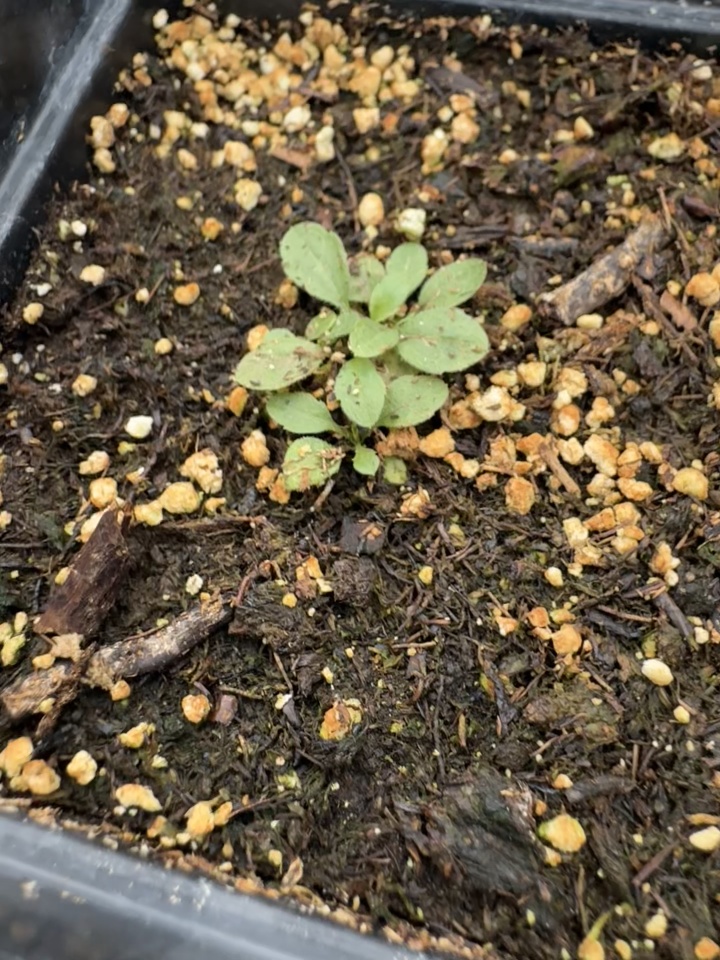}\hspace{\hspacing}
    \includegraphics[width=\growthfigsize\textwidth]{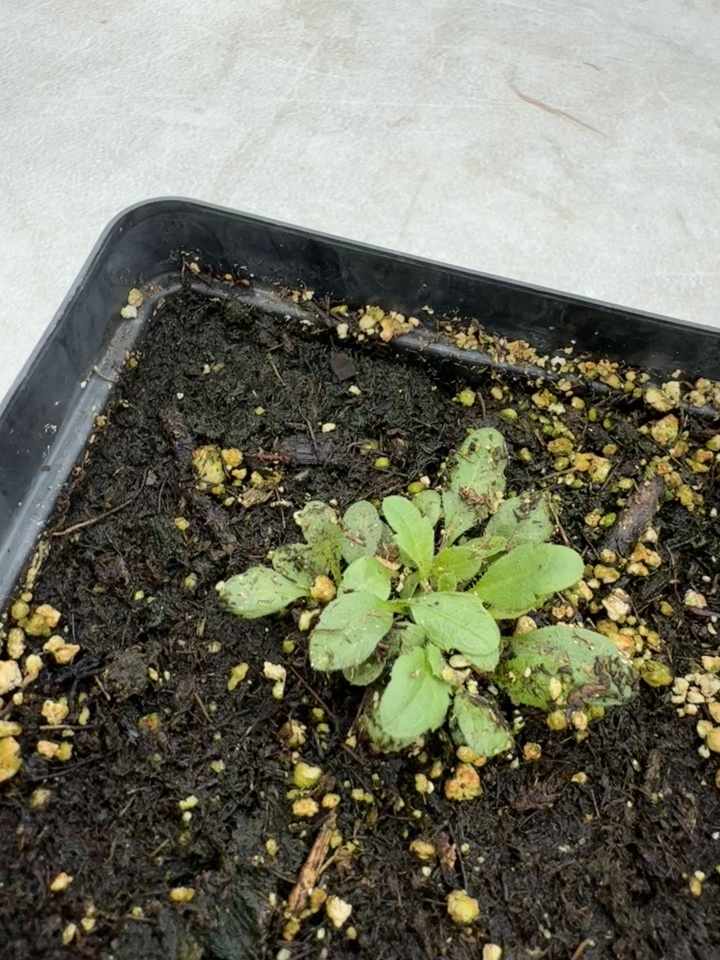}\hspace{\hspacing}
    \includegraphics[width=\growthfigsize\textwidth]{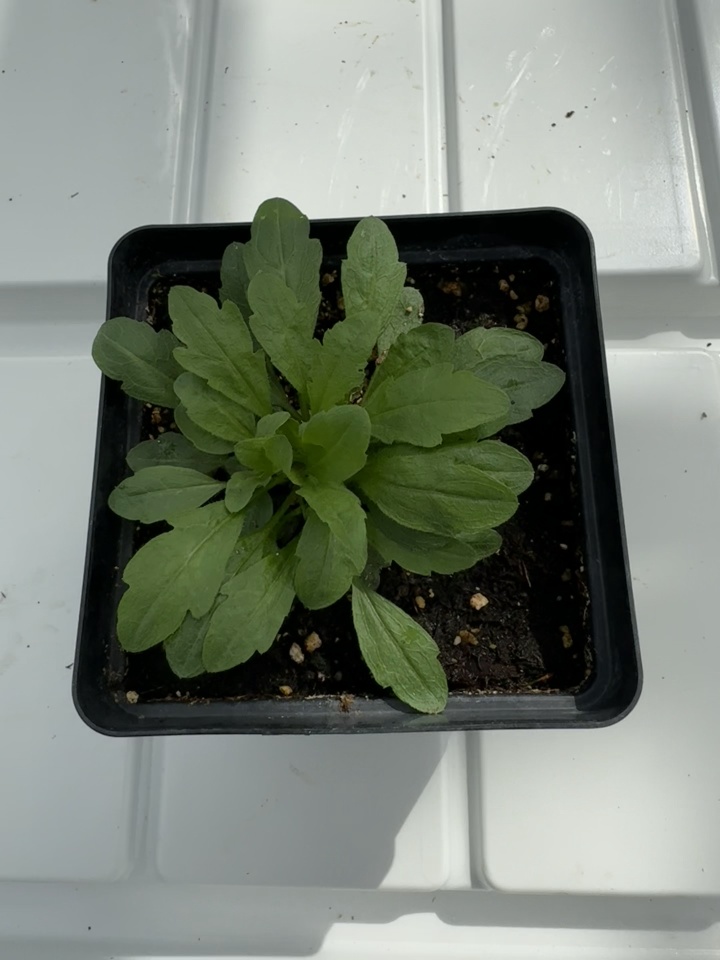}\hspace{\hspacing}
    \includegraphics[width=\growthfigsize\textwidth]{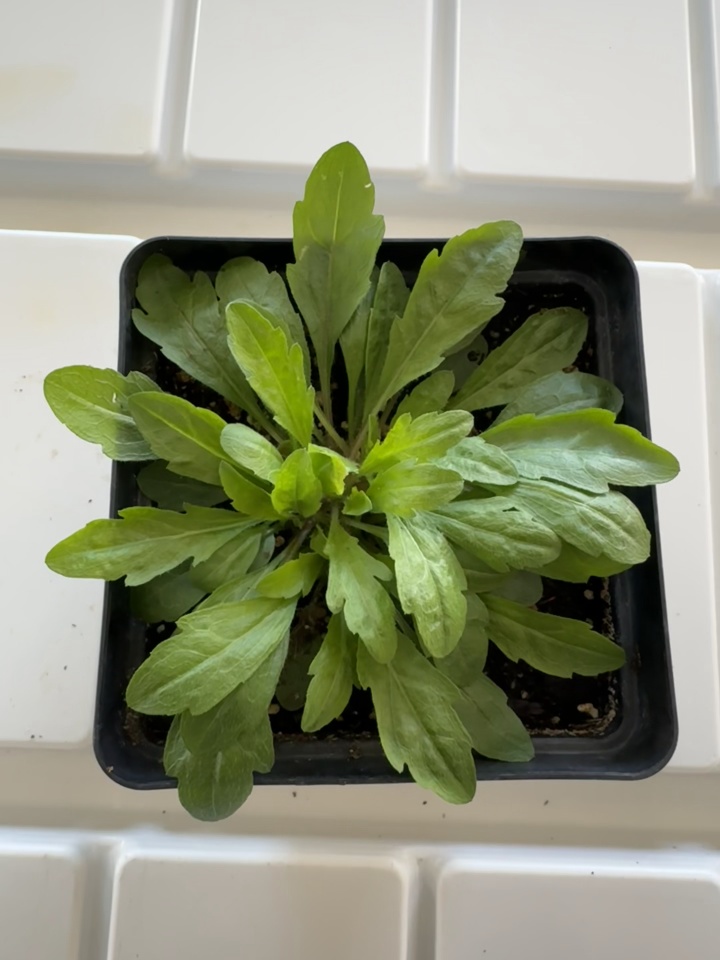}\hspace{\hspacing}
    \includegraphics[width=\growthfigsize\textwidth]{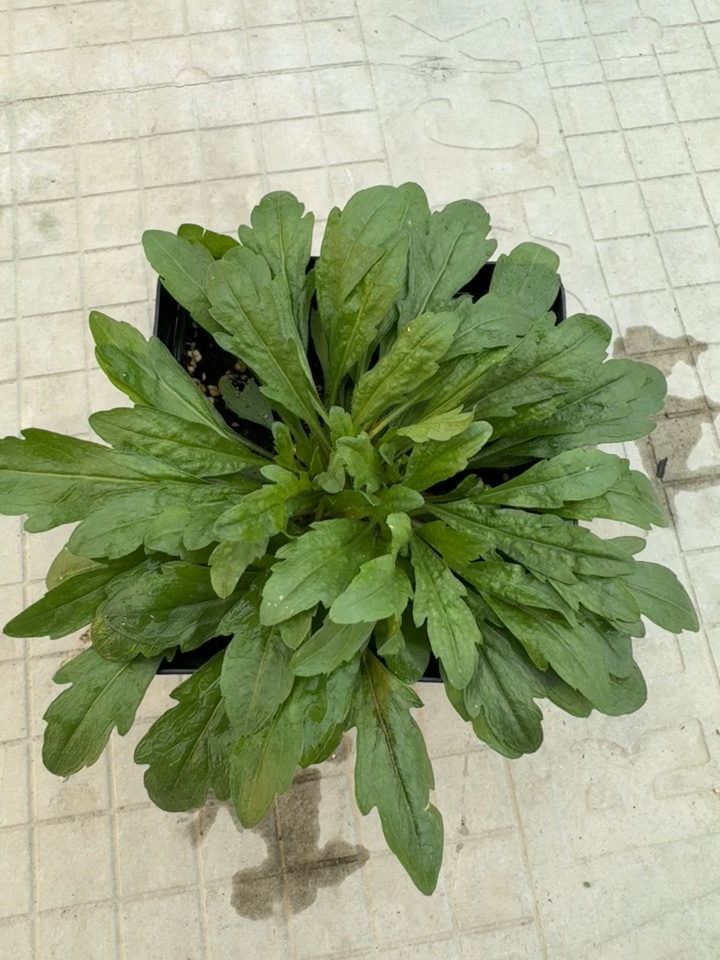}\hspace{\hspacing}
    \includegraphics[width=\growthfigsize\textwidth]{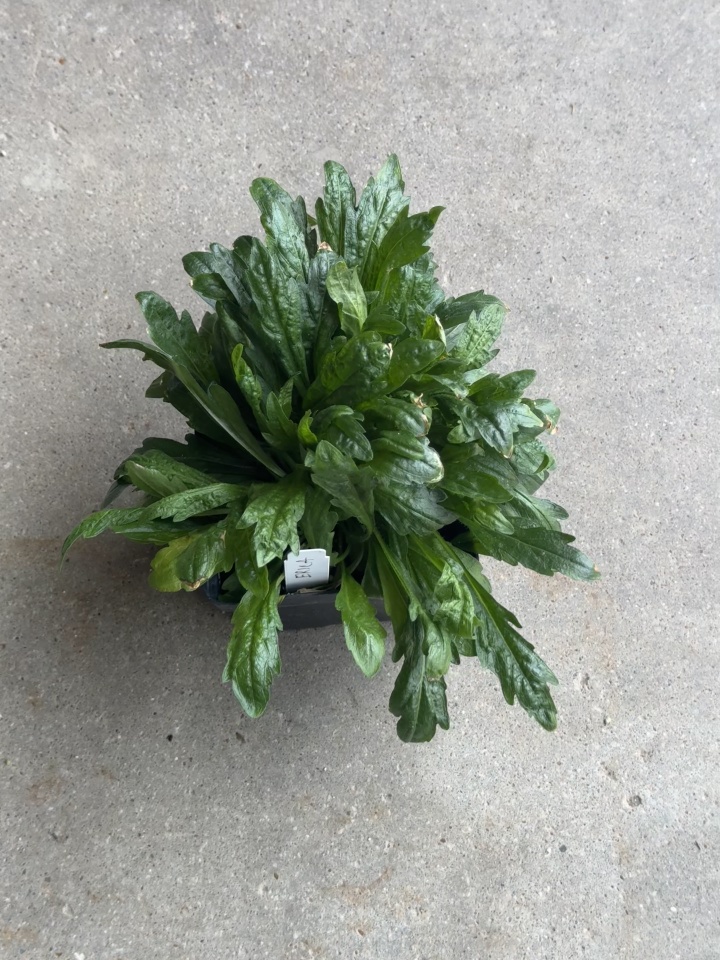}\hspace{\hspacing}
    \includegraphics[width=\growthfigsize\textwidth]{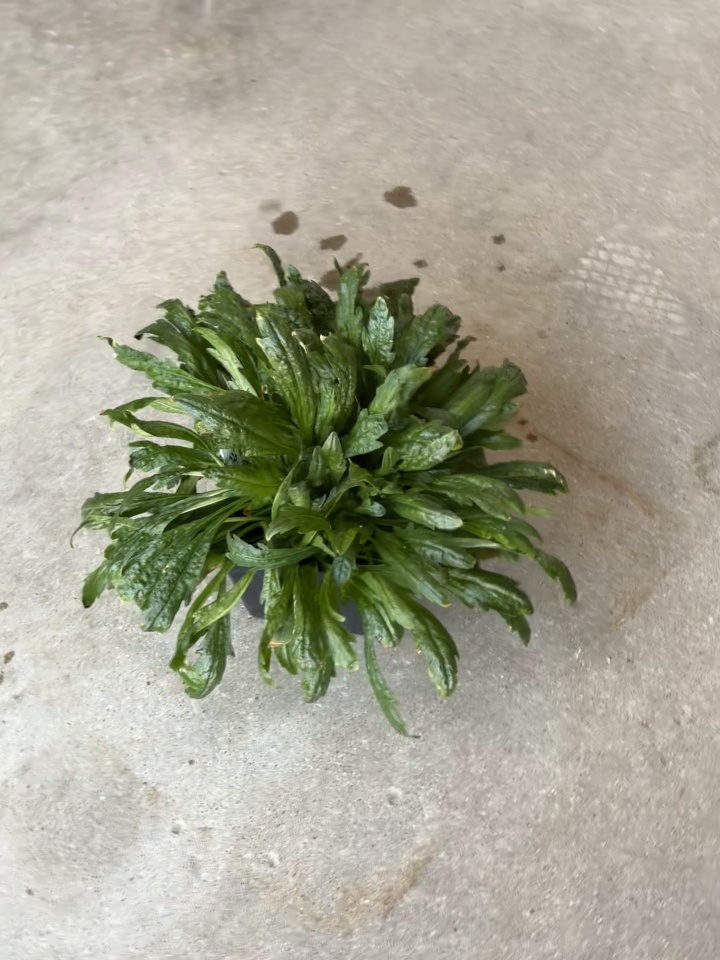}
    \\[0pt]

    \caption{Growth progression over 11 weeks for three subjects (SETFA, AMARE, and ERICA), showing weekly development from week 1 to week 11. Best viewed on screen.}
    \label{fig:growth-progression}
    \vspace{-15pt}
\end{figure*}

We present a novel multi-task temporal dataset of 16 weed species growth patterns for semantic segmentation, height regression, and growth stage classification. The dataset spans 11 weeks from sprouting through flowering, providing comprehensive coverage of the primary growth cycle for weed species commonly found in Midwestern cropping systems of the USA.

\begin{table}[th]
% \footnotesize
\scriptsize
  \centering
  \captionsetup{belowskip=5pt}
  \setlength{\tabcolsep}{1pt}  % Reduce column spacing
  \begin{tabular}{lcccc}
    \toprule
     \multirow{2}{*}{\makecell{Species~\cite{kotleba1994european} (Scientific Name~\cite{borsch2020world})}} & \makecell{Image} & \multicolumn{2}{c}{Height (cm)} & Growth \\
    % \cmidrule(lr){3-4}
     & Count & \makecell{Max} & \makecell{Std Dev} & \makecell{(cm/week)} \\
    
    \midrule
    % \rowcolor[gray]{0.95} \multicolumn{5}{l}{\textit{Fast-growing species ($>$10 cm/week)}} \\
    \rowcolor{gray!20}
    \multicolumn{5}{l}{\textit{Fast-growing species ($>$10 cm/week)}} \\
    \midrule
    AMATU (Amaranthus tuberculatus (Moq.) Sauer.) & 7,457 & 155.00 & 46.34 & 13.72 \\
    SORHA (Sorghum halepense (L.) Pers.) & 5,540 & 121.00 & 45.57 & 14.06 \\
    SETFA (Setaria faberi Herrm.) & 7,787 & 124.00 & 42.39 & 11.75 \\
    % \rowcolor[gray]{0.95} \multicolumn{5}{l}{\textit{Medium-growing species (5-10 cm/week)}} \\
    \midrule
    \rowcolor{gray!20}
    \multicolumn{5}{l}{\textit{Medium-growing species (5-10 cm/week)}} \\
    \midrule
    SORVU (Sorghum bicolor (L.) Moench.) & 4,787 & 100.00 & 38.21 & 9.84 \\
    PANDI (Panicum dichotomiflorum Michx.) & 7,577 & 87.00 & 32.46 & 8.40 \\
    SETPU (Setaria pumila (Poir.) Roem.) & 7,971 & 99.00 & 29.98 & 8.20 \\
    DIGSA (Digitaria sanguinalis (L.) Scop.) & 9,211 & 77.00 & 27.84 & 7.53 \\
    ECHCG (Echinochloa crus-galli (L.) P. Beauv.) & 8,562 & 80.00 & 26.35 & 7.38 \\
    SIDSP (Sida spinosa L.) & 6,977 & 69.00 & 24.10 & 6.77 \\
    AMARE (Amaranthus retroflexus L.) & 7,951 & 75.00 & 23.83 & 6.86 \\
    ABUTH (Abutilon theophrasti Medik.) & 8,770 & 72.00 & 22.55 & 6.32 \\
    AMBEL (Ambrosia artemisiifolia L.) & 8,630 & 71.00 & 22.07 & 6.19 \\
    % \rowcolor[gray]{0.95} \multicolumn{5}{l}{\textit{Slow-growing species ($<$5 cm/week)}} \\
    \midrule
    \rowcolor{gray!20}
    \multicolumn{5}{l}{\textit{Slow-growing species ($<$5 cm/week)}} \\
    \midrule
    AMAPA (Amaranthus palmeri S. Watson.) & 9,080 & 62.00 & 19.82 & 5.66 \\
    CYPES (Cyperus esculentus L.) & 8,131 & 56.00 & 18.26 & 5.42 \\
    CHEAL (Chenopodium album L.) & 4,670 & 30.00 & 12.97 & 2.86 \\
    ERICA (Erigeron canadensis L.) & 7,240 & 17.30 & 6.37 & 1.70 \\
    \bottomrule
  \end{tabular}
  \vspace{-5pt}
  \caption{Statistical summary of our weed species dataset (120,341 images) categorized by growth rates. The data shows significant variation in maximum height (17.3--155.0 cm) and growth patterns (1.70--14.06 cm/week), presenting diverse challenges for our multi-task learning approach.}
  \label{tab:species_stats}
  \vspace{-15pt}
\end{table}

\noindent \textbf{Data Collection. } We conducted this study during the spring and summer of 2024 at the SIU Horticulture Research Center greenhouse facility. 
The greenhouse was equipped with 1000W High Pressure Sodium grow lights maintaining optimal temperatures of 30--32$^\circ$C.
We used 32 square containers (10.7 cm × 10.7 cm × 9 cm) with two containers per species, where each container housed a single weed plant, creating replicate pairs for each species. Each container was filled with Pro-Mix® BX potting medium.
Weeds were maintained through regular watering based on soil moisture requirements and an all-purpose 20--20--20 nutrient solution applied every three days. Data acquisition utilized an iPhone 15 Pro Max positioned 1.5 feet above specimens, capturing 360-degree video documentation at 1440 × 1920 resolution and 30 FPS. We recorded 349 videos (15--30 seconds each) on a weekly basis throughout developmental stages, with the camera rotating around each plant to capture all viewing angles. Video recording was chosen over still images to ensure comprehensive angular coverage of each specimen in a single capture session.
Minor protocol deviations occurred during collection: one plant each from SORVU and CHEAL duplicate pairs died after the third week, and SORHA weeds emerged only after the second week rather than the planned first week, while all remaining species successfully emerged within the first week. Despite these variations, consistent data collection procedures were maintained across all species throughout the complete 11-week monitoring period.
We implemented systematic growth stage monitoring based on established phenological standards, with weekly imaging from week 1 (BBCH stage 11, first true leaf visible) through week 11 (BBCH stage 60, initial flower appearance). This temporal framework captured the complete vegetative growth cycle and transition to reproductive development across all species. \cref{fig:growth-progression} illustrates the temporal progression across three representative species (SETFA, AMARE, ERICA), demonstrating the diverse growth patterns and morphological changes captured in our dataset.

\noindent \textbf{Data Preprocessing and Annotation.} We applied two preprocessing steps: (1) temporal downsampling by extracting every 2nd frame from the 30 FPS videos to reduce redundancy, and (2) spatial downscaling to 720 × 960 pixels while preserving the original 3:4 aspect ratio. The resulting dataset contains 120,341 frames extracted from 349 videos, divided into training (80\%), validation (10\%), and test (10\%) sets using frame-level splitting ensuring balanced species representation across splits.

For annotation, we employed the SAM2-Hiera-L~\cite{ravi2024sam2} model to semi-automatically generate segmentation masks, with manual verification and correction. Plant components including stems, leaves, and flowers are labeled as a single foreground category. Each image is annotated with its corresponding growth week (1-11) based on capture date. We manually measured weed heights on a weekly basis, recording 325 measurements that revealed substantial variation across species, from 0.2 cm to 155 cm ($\Delta$ = 154.8 cm). 
Table~\ref{tab:species_stats} presents key statistics for each species, showing significant inter- and intra-species variability. AMATU demonstrates the most aggressive growth, averaging 13.72 cm weekly increase and reaching 155 cm, while ERICA exhibits the slowest growth (1.70 cm/week) with a maximum height of just 17.3 cm. The height data reveals substantial intra-species variability, with standard deviations ranging from 6.37 cm to 46.34 cm, presenting challenging regression targets for our models. The resulting dataset combines high-resolution RGB images, segmentation masks, height measurements, and weekly growth stage labels, enabling comprehensive analysis of weed species growth patterns throughout their complete life cycle.

\vspace{-5pt}
\section{Method}%
\label{sec:architecture}
% \vspace{-5pt}

This section introduces our Dual-path UIB Encoder with Multi-Task Bifurcated Decoder, an efficient framework for semantic segmentation and temporal growth analysis. As shown in Fig.~\ref{fig:architecture}, our model consists of two main encoder pathways that extract complementary information from input RGB images, and a bifurcated decoder that enables joint learning of multiple tasks. We denote feature maps as $C \times H \times W$ where $C$, $H$, and $W$ represent spatial channel dimensions, height, and width respectively.

\subsection{Dual-path UIB Encoder}
\label{sec:dual-path-encoder}

Building upon the dual-path design of BiSeNetV2~\cite{yu2021bisenet}, our encoder efficiently balances spatial detail preservation and semantic context extraction through specialized branches while incorporating UIB blocks for enhanced feature representation.

\noindent \textbf{Detail Branch.} The Detail Branch captures fine-grained spatial details crucial for accurate boundary delineation. Following a shallow-wide architecture inspired by VGGNet~\cite{simonyan2014very}, it consists of three sequential stages (S1, S2, S3) that generate feature maps with progressively reduced spatial resolutions ($H/2$, $H/4$, $H/8$) and expanded channel dimensions (64, 64, 128). Each stage applies 3$\times$3 convolutions followed by batch normalization and ReLU activation.

\noindent \textbf{Semantic Branch.} The Semantic Branch follows a deep-narrow architecture with aggressive downsampling to efficiently capture semantic context. It begins with a Stem Block for efficient initial feature extraction and downsampling, inspired by Inception~\cite{szegedy2015going} networks to balance computational efficiency with feature richness. The branch then employs hierarchical UIB blocks organized in three stages (S3, S4, S5\_1) that progressively downsample features while increasing channel capacity, concluding with a context embedding block for global context enhancement.

\noindent \textbf{Universal Inverted Bottleneck Blocks.} As the core component of our Semantic Branch, we replace the original Gather-and-Expansion blocks~\cite{yu2021bisenet} from BiSeNetV2 with UIB blocks from MobileNetV4~\cite{qin2024mobilenetv4}. UIB blocks use configuration notation S[start]-M[mid]-E[end] representing kernel sizes for start, middle, and end depthwise convolutions, where kernel size 0 indicates the operation is skipped (identity mapping). Our S0-M3-E0 configuration employs only the middle 3×3 depthwise convolution:

{\footnotesize
\vspace{-5pt}
\begin{equation}
\begin{aligned}
F_{exp} &= \text{Conv}_{1\times1}(F_{in}) \cdot \text{ExpRatio} \\
F_{dw} &= \text{DWConv}_{3\times3}(F_{exp}) \\
F_{se} &= \text{SE}(F_{dw}) \\
F_{out} &= \text{LayerScale}(\text{Conv}_{1\times1}(F_{se})) + F_{in} \cdot \delta
\end{aligned}
\end{equation}
\vspace{-5pt}
}

\noindent where ExpRatio=6, SE enables adaptive channel recalibration with reduction ratio 0.25, LayerScale ensures stable training~\cite{touvron2021going}, and $\delta$ enables residual connections when dimensions match. The UIB blocks are organized hierarchically across three stages (S3, S4, S5\_1) with progressive downsampling and channel expansion. To enhance global context modeling, we apply a Context Embedding Block after the final UIB stage that captures global statistical information through adaptive average pooling and residual connections.

\begin{figure}[t] \centering
    \makebox[0.108\textwidth]{\scriptsize Image}
    \makebox[0.108\textwidth]{\scriptsize Segmentation}
    \makebox[0.108\textwidth]{\scriptsize Height}
    \makebox[0.108\textwidth]{\scriptsize Growth Stage}
    \\
    \includegraphics[width=0.108\textwidth]{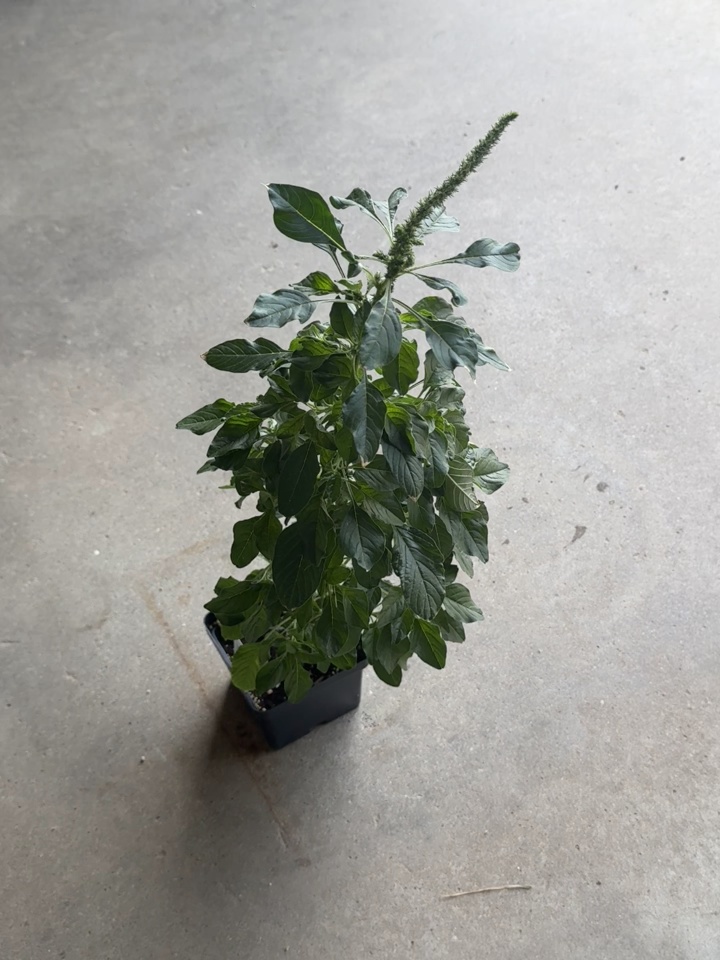}
    \includegraphics[width=0.108\textwidth]{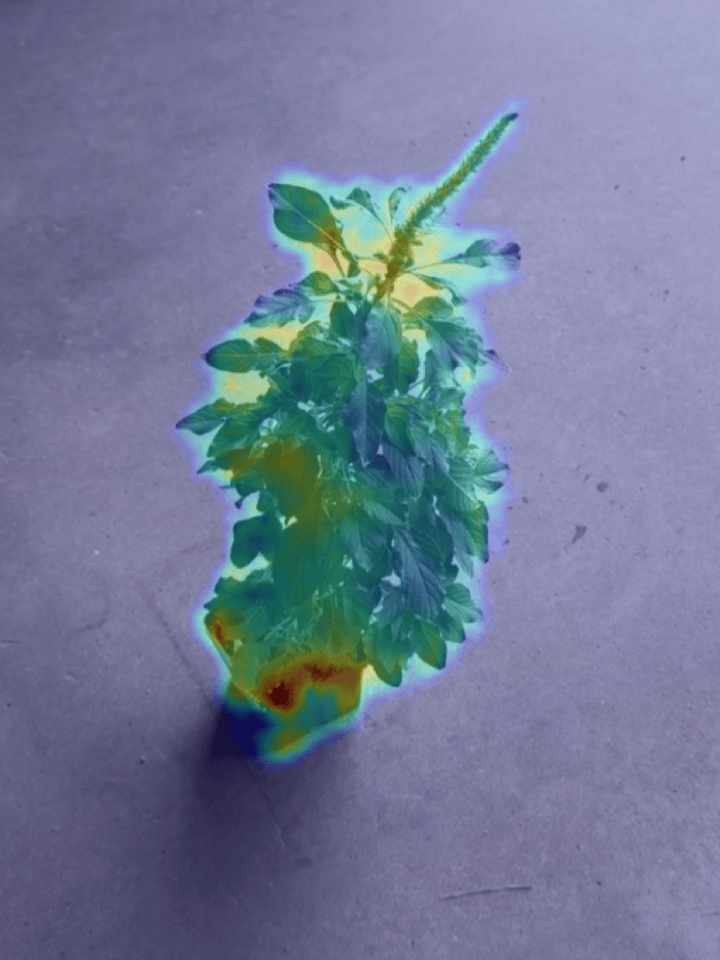}
    \includegraphics[width=0.108\textwidth]{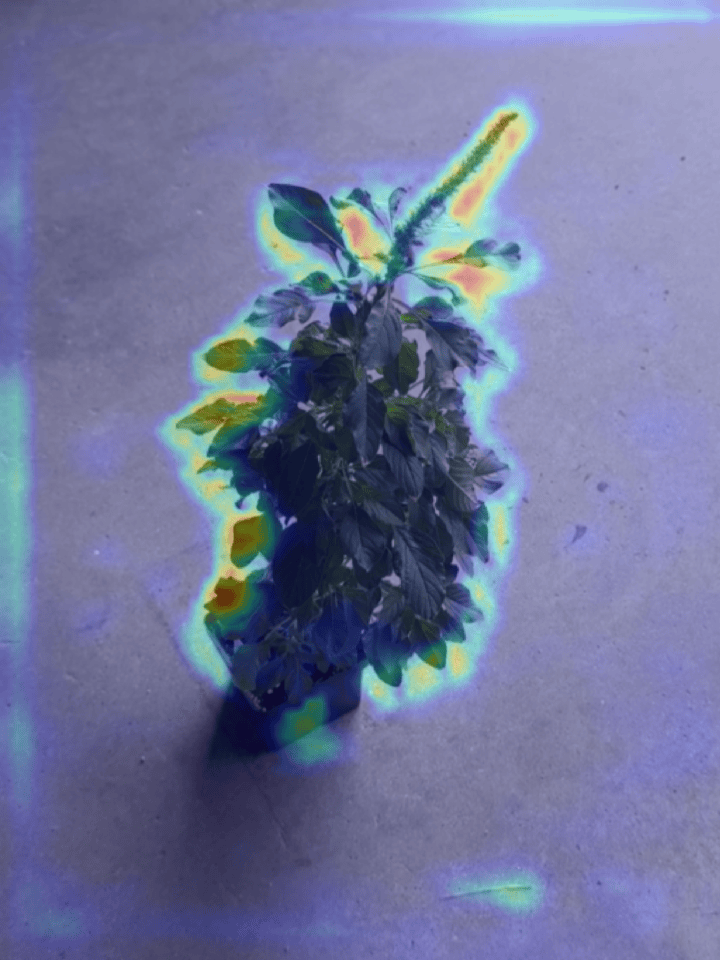}
    \includegraphics[width=0.108\textwidth]{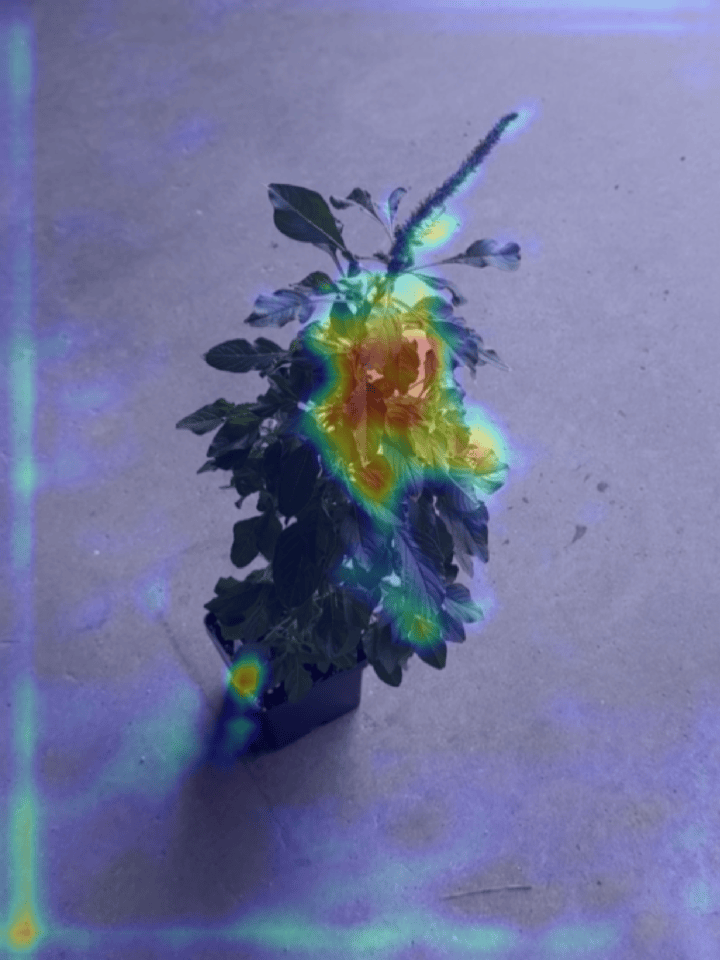}
    \\
    \caption{Visualization of attention activation maps from the aggregation layer for AMAPA at week 10. Each task displays distinct activation patterns: segmentation shows uniform boundary-focused activation, height estimation concentrates on plant extremities, and growth stage classification focuses on stem and mature leaf regions containing temporal growth indicators.}
  \label{fig:activation-maps}
  \vspace{-15pt}
\end{figure}

\begin{table*}[t]
\centering
\resizebox{\textwidth}{!}{
\scriptsize
\begin{tabular}{l|cc|ccccccc|cc}
\toprule
& \multicolumn{2}{c|}{Segmentation} & \multicolumn{7}{c|}{Height Estimation} & \multicolumn{2}{c}{Growth Stage Classification} \\
\cmidrule(lr){2-3} \cmidrule(lr){4-10} \cmidrule(lr){11-12}
Model & \makecell{mIoU\\(\%)$\uparrow$} & \makecell{mF1\\(\%)$\uparrow$} & \makecell{MAE\\(cm)$\downarrow$} & \makecell{RMSE\\(cm)$\downarrow$} & \makecell{R$^2$\\$\uparrow$} & \makecell{Max Error\\(cm)$\downarrow$} & \makecell{Within\\1cm (\%)$\uparrow$} & \makecell{Within\\2cm (\%)$\uparrow$} & \makecell{Within\\5cm (\%)$\uparrow$} & \makecell{Accuracy\\(\%)$\uparrow$} & \makecell{F1\\(\%)$\uparrow$} \\
\midrule
MTL-SegFormer & 56.17 & 70.56 & 5.10 & 8.03 & 0.9289 & 61.07 & 25.21 & 40.15 & 65.16 & 97.96 & 98.22 \\
MTL-UNet & 51.51 & 66.3 & 2.15 & 3.03 & 0.9899 & 29.22 & 39.07 & 60.5 & 89.14 & 99.97 & 99.97 \\
MTL-Poolformer & 72.45 & 83.53 & 4.20 & 6.74 & 0.9499 & 52.52 & 29.08 & 45.29 & 72.41 & 98.70 & 98.89 \\
MTL-BiSeNetV1 & 87.25 & 93.08 & 1.97 & 2.81 & 0.9913 & \textbf{18.89} & 41.22 & 63.90 & 91.51 & 99.99 & 99.99 \\
MTL-BiSeNetV2 & 89.29 & 94.26 & 1.75 & 2.51 & 0.9930 & 24.39 & \textbf{44.11} & 68.13 & 94.30 & \textbf{100.00} & \textbf{100.00} \\
MTL-SFNet & 86.27 & 92.51 & 4.21 & 6.66 & 0.9511 & 69.24 & 26.36 & 44.53 & 72.33 & 98.45 & 98.74 \\
\midrule
\rowcolor{gray!20}
WeedSense & \textbf{89.78} & \textbf{94.54} & \textbf{1.67} & \textbf{2.32} & \textbf{0.9941} & 19.34 & 43.26 & \textbf{70.37} & \textbf{95.49} & 99.99 & 99.99 \\
\bottomrule
\end{tabular}
}
\vspace{-5pt}
\caption{Quantitative comparison of multi-task learning models for weed species analysis. Our WeedSense model achieves best or competitive performance across segmentation, height estimation, and growth stage classification metrics. Bold indicates best results.}
\label{tab:performance_comparison}
\vspace{-15pt}
\end{table*}

\noindent \textbf{Aggregation Layer.} Having extracted complementary information through our dual-path design, we now need to effectively combine these heterogeneous features. The Aggregation Layer uses semantic features as intelligent guides to direct where detail features should focus, creating a unified representation that preserves spatial precision while incorporating semantic understanding. This aggregated representation then serves as input to our multi-task decoder.

\noindent \textbf{Auxiliary Supervision Strategy.} We implement four auxiliary segmentation heads connected to different stages of the Semantic Branch (Stem Block, S3, S4, and S5\_1), with stage-specific upsampling factors (4$\times$, 8$\times$, 16$\times$, and 32$\times$ respectively). These heads facilitate effective gradient flow during training while being discarded during inference, thus incurring no additional computational cost.

%
% \vspace{-5pt}
\subsection{Multi-Task Bifurcated Decoder}%

Our MTBD processes the aggregated features ($128 \times H/8 \times W/8$) from the encoder through parallel pathways for semantic segmentation and temporal growth prediction.

\noindent \textbf{Semantic Segmentation Head.} The segmentation head follows an encoder-decoder structure with progressive upsampling to recover full spatial resolution:

{\footnotesize
\vspace{-5pt}
\begin{equation}
\begin{aligned}
F_{mid} &= \text{Dropout}(\text{Conv}_{3\times3}(F_{agg})) \\
M &= \text{PixelShuffle}(\text{Conv}_{1\times1}(F_{mid}))
\end{aligned}
\end{equation}
\vspace{-5pt}
}

\noindent where $F_{agg}$ represents the aggregated features and $M$ is the predicted segmentation mask at resolution $N_{cls} \times H \times W$, with $N_{cls}$ representing the number of classes (17 classes including background). PixelShuffle~\cite{shi2016real} provides parameter-free 8$\times$ upsampling by reorganizing channel data into spatial dimensions.

\noindent \textbf{Temporal Growth Decoder.} The TGD predicts plant height and growth stage through global feature processing. Aggregated features undergo adaptive average pooling to produce a 128-dimensional representation, followed by linear projection to 512-dimensional embeddings. The projected features are processed by a transformer block with multi-head self-attention and layer normalization:

{\footnotesize
\vspace{-5pt}
\begin{equation}
\begin{aligned}
F_{attn} &= \text{LayerNorm}(F_{proj} + \text{MHA}(F_{proj}, F_{proj}, F_{proj})) \\
F_{trans} &= \text{LayerNorm}(F_{attn} + \text{FFN}(F_{attn}))
\end{aligned}
\end{equation}
\vspace{-5pt}
}

\noindent where MHA represents multi-head attention (8 heads) and FFN is a two-layer feed-forward network (512$\rightarrow$2048$\rightarrow$512 dimensions) with GELU activation. Finally, two parallel task-specific heads process the transformed features:

{\footnotesize
  \vspace{-5pt}
  \begin{equation}
  \begin{aligned}
  F_{\text{task}} &= \text{ReLU}(\text{LayerNorm}(\text{Linear}_{512}(\text{Linear}_{1024}(F_{\text{trans}})))) \\
  \hat{h} &= \text{Linear}_1(F_{\text{task}}) \\
  \hat{w} &= \text{Linear}_{11}(F_{\text{task}})
  \end{aligned}
  \end{equation}
  \vspace{-5pt}
  }

\noindent where $\hat{h}$ represents plant height prediction (in cm) and $\hat{w}$ represents growth stage classification logits using weekly intervals (weeks 1-11). The architecture employs hard parameter sharing in the feature processing pipeline while maintaining separate head weights, facilitating knowledge transfer between related tasks~\cite{ruder2017overview}.

\noindent \textbf{Feature Activation Visualization.} \Cref{fig:activation-maps} shows Grad-CAM~\cite{selvaraju2020grad} activation maps from our aggregation layer for AMAPA at week 10. The maps demonstrate task-specific feature extraction. Segmentation shows uniform activation across plant boundaries. This helps with accurate boundary delineation. Height estimation focuses on plant extremities and uppermost leaves. These regions correspond to vertical measurements. Growth stage classification targets the central stem and mature leaf regions. These areas contain visual features like stem thickness and leaf development that indicate developmental progression through weekly intervals. The distinct patterns confirm our multi-task architecture learns complementary representations. There is no interference between tasks. This enables high performance across all tasks while sharing network parameters.

\vspace{-5pt}
\section{Experiments and Discussion}%
\label{sec:results}
% \vspace{-5pt}

\subsection{Implementation Details}%
\label{sec:implementation_details}
All experiments are conducted using PyTorch on an Intel Xeon Gold 6240 CPU @ 2.60GHz system with 124 GB RAM and an NVIDIA Tesla V100S GPU with 32GB memory. We evaluate our approach against six state-of-the-art architectures adapted for multi-task learning: SegFormer~\cite{xie2021segformer}, UNet~\cite{ronneberger2015u}, PoolFormer~\cite{yu2022metaformer}, BiSeNetV1~\cite{yu2018bisenet}, BiSeNetV2~\cite{yu2021bisenet}, and SFNet~\cite{lee2019sfnet}. For fair comparison, each method uses its original encoder-decoder architecture for semantic segmentation while incorporating our proposed Temporal Growth Decoder (TGD) for height regression and growth stage classification tasks. 
The model is trained using a multi-task loss function that equally weights three components: pixel-level weighted cross-entropy for segmentation (including auxiliary supervision), Mean Squared Error for height regression, and cross-entropy for growth stage classification. 
All models are trained for 50 epochs from scratch without pre-trained weights. Training uses the Adam optimizer with initial learning rate $\eta_{\text{base}} = 2 \times 10^{-4}$, weight decay of 0.0001, and cosine annealing schedule~\cite{loshchilov2017decoupled}. A 1,500-iteration linear warmup period gradually increases the learning rate from $0.1\eta_{\text{base}}$ to $\eta_{\text{base}}$.
Data augmentation includes random cropping (50\%-200\% of original size), horizontal flipping (50\% probability), and ImageNet normalization. Random cropping is applied to all training images, while horizontal flipping is applied probabilistically to half the samples during each epoch. All images are resized to $512 \times 512$ pixels with batch size 8 during training and batch size 1 during evaluation.

\begin{figure}[t]
\vspace{-15pt}
    \centering
    \includegraphics[width=0.45\textwidth]{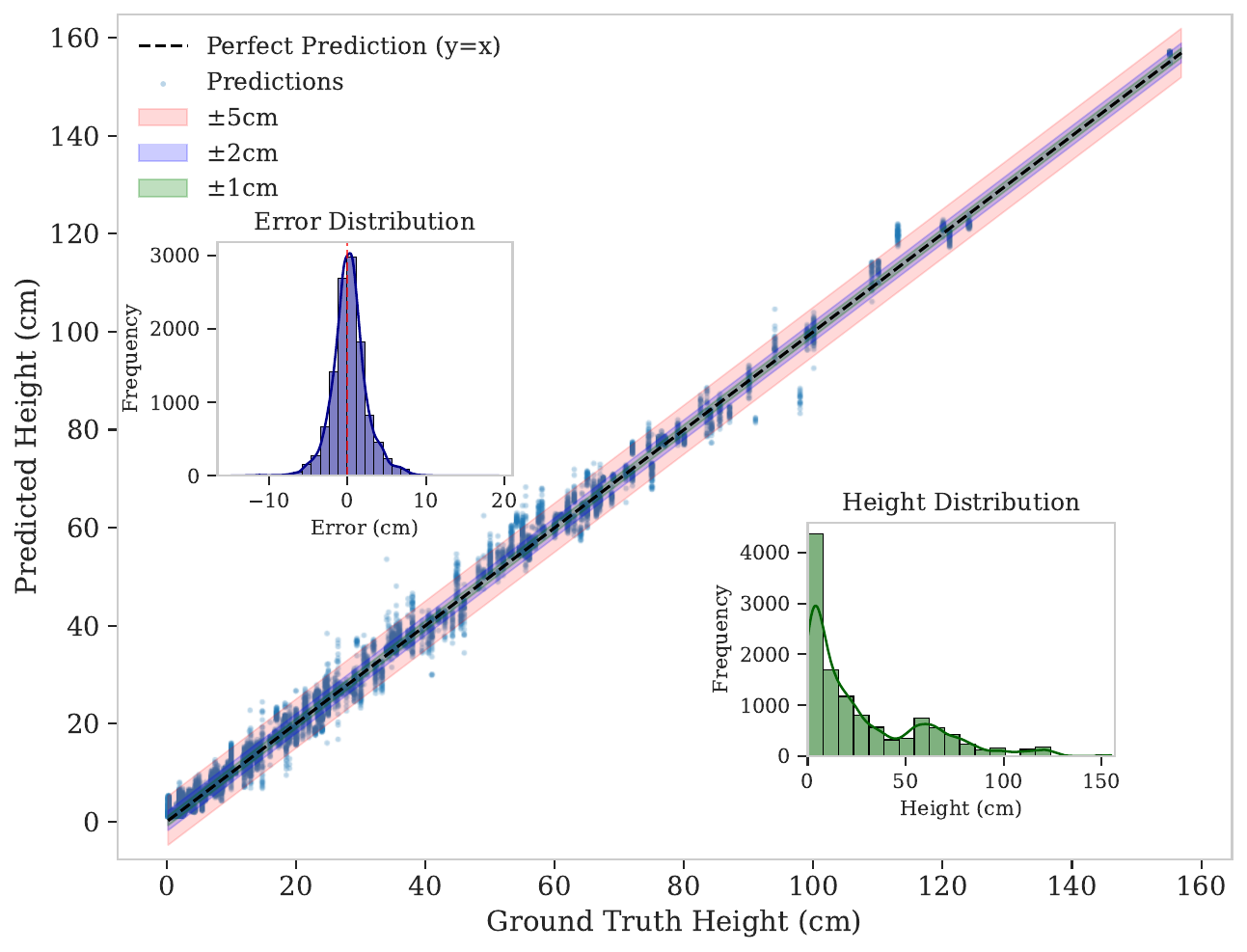}
    \vspace{-5pt}
    \caption{Height estimation performance of WeedSense across the full measurement range (0.2-155 cm). Predictions cluster tightly along the perfect prediction line (y=x) with tolerance bands showing ±1cm (green), ±2cm (blue), and ±5cm (red) accuracy zones. The error distribution (top left inset) is symmetric around zero with no systematic bias, while the height distribution (bottom right inset) shows the diverse range of plant sizes in our dataset.}
% \vspace{-5pt}
    \label{fig:height_regression}
    \vspace{-10pt}
\end{figure}

% \vspace{-5pt}
\subsection{Comparative Evaluation}%
\label{sec:comparative_evaluation}
% \vspace{-5pt}

We evaluate our proposed WeedSense against the multi-task learning variants we created: MTL-SegFormer, MTL-UNet, MTL-PoolFormer, MTL-BiSeNetV1, MTL-BiSeNetV2, and MTL-SFNet. 
~\Cref{tab:performance_comparison,tab:height_by_size,tab:comp_efficiency} summarize results across segmentation quality, height estimation, growth stage classification, and computational efficiency.

\noindent \textbf{Segmentation Performance.} As shown in ~\Cref{tab:performance_comparison}, WeedSense achieves the best segmentation performance with 89.78\% mIoU and 94.54\% mF1 score. Among competitive methods, WeedSense surpasses MTL-BiSeNetV2 by 0.49 and 0.28 percentage points respectively, outperforms MTL-BiSeNetV1 by 2.53 percentage points, and exceeds MTL-SFNet by 3.51 percentage points in mIoU.
WeedSense achieves significant gains over lower-performing methods: 38.27 percentage points better than MTL-UNet, 33.61 percentage points better than MTL-SegFormer, and 17.33 percentage points better than MTL-PoolFormer. The qualitative results in ~\cref{fig:qualitative-comparison} further validate these findings, showing that WeedSense provides more accurate boundary delineation for AMATU, CYPES, ECHCG species with better precision for fine-grained plant structures.

\noindent \textbf{Height Estimation Performance.} Height regression presents challenges due to the wide range of plant heights from 0.2 to 155 cm and visual occlusions during flowering stages. As detailed in ~\Cref{tab:performance_comparison}, WeedSense achieves the best overall performance with 1.67 cm MAE, 2.32 cm RMSE, and 0.9941 R² value.
Our approach achieves a maximum error of 19.34 cm versus the best competitor's 24.39 cm, representing a 20.7\% reduction. For tolerance rates, WeedSense achieves 43.26\% of predictions within 1 cm, 70.37\% within 2 cm, and 95.49\% within 5 cm. The 2 cm tolerance rate surpasses all competitors by 2.24-30.22 percentage points.

~\Cref{tab:height_by_size} reveals consistent performance across plant sizes. For small plants (0-20 cm), WeedSense matches the best performer at 1.20 cm error. For medium plants (20-50 cm), our approach achieves 2.28 cm error, representing a 9.9\% improvement over the best competitor. For large (50-100 cm) and very large plants ($>$100 cm), WeedSense maintains 2.28 cm and 2.60 cm errors respectively, outperforming all other methods. ~\Cref{fig:height_regression} visualizes predictions clustering tightly along the identity line with consistent accuracy across all height ranges.

\begin{table}[t]
\resizebox{0.48\textwidth}{!}{
\centering
\scriptsize
\begin{tabular}{lcccc}  % Removed @{} to fix rowcolor issue
\toprule
\multirow{2}{*}{Model} & \makecell{Small$\downarrow$\\0-20cm} & \makecell{Medium$\downarrow$\\20-50cm} & \makecell{Large$\downarrow$\\50-100cm} & \makecell{Very Large$\downarrow$\\$>$100cm} \\
\midrule
MTL-SegFormer & 2.81 cm & 7.50 cm & 8.18 cm & 11.29 cm \\
MTL-UNet & 1.46 cm & 3.03 cm & 3.10 cm & 3.01 cm \\
MTL-Poolformer & 2.23 cm & 6.11 cm & 7.10 cm & 8.89 cm \\
MTL-BiSeNetV1 & 1.35 cm & 2.60 cm & 2.96 cm & 2.74 cm \\
MTL-BiSeNetV2 & \textbf{1.20 cm} & 2.53 cm & 2.32 cm & 3.12 cm \\
MTL-SFNet & 2.12 cm & 5.64 cm & 7.65 cm & 10.22 cm \\
\midrule
\rowcolor{gray!20}
WeedSense & \textbf{1.20 cm} & \textbf{2.28 cm} & \textbf{2.28 cm} & \textbf{2.60 cm} \\
\bottomrule
\end{tabular}
}
\vspace{-5pt}
\caption{Height estimation error (MAE) across different plant size categories.}
\label{tab:height_by_size}
\end{table}

\begin{table}[t]
\vspace{-5pt}
\centering
\resizebox{0.4\textwidth}{!}{

\scriptsize
% \footnotesize
% \setlength{\tabcolsep}{3pt}  % Adjust column spacing
\begin{tabular}{lccc}
\toprule
Model & Params (M)$\downarrow$ & GFLOPs$\downarrow$ & FPS$\uparrow$ \\
\midrule
MTL-SegFormer & \textbf{8.19} & \textbf{7.94} & 138 \\
MTL-UNet & 35.67 & 233.23 & 94 \\
MTL-Poolformer & 18.77 & 22.86 & 101 \\
MTL-BiSeNetV1 & 17.31 & 13.30 & \textbf{249} \\
MTL-BiSeNetV2 & 29.62 & 16.84 & 185 \\
MTL-SFNet & 18.62 & 30.77 & 151 \\
\midrule
\rowcolor{gray!20}
WeedSense & 30.50 & 16.73 & 160 \\
\bottomrule
\end{tabular}
}
\vspace{-5pt}
\caption{Computational efficiency comparison showing model size, computational complexity, and inference speed.}
\label{tab:comp_efficiency}
\vspace{-10pt}
\end{table}

% Define sizing commands for qualitative comparison
\newcommand{\qualfigsize}{0.10}
\newcommand{\quallabelwidth}{0.02}
\newcommand{\qualhspacing}{-1.5pt}

\begin{figure*}[t] 
    \centering
    
    % Method labels header
    \makebox[\quallabelwidth\textwidth][c]{\scriptsize}
    \makebox[\qualfigsize\textwidth][c]{\scriptsize Input Image}\hspace{\qualhspacing}
    \makebox[\qualfigsize\textwidth][c]{\scriptsize Ground Truth}\hspace{\qualhspacing}
    \makebox[\qualfigsize\textwidth][c]{\scriptsize MTL-SegFormer}\hspace{\qualhspacing}
    \makebox[\qualfigsize\textwidth][c]{\scriptsize MTL-UNet}\hspace{\qualhspacing}
    \makebox[\qualfigsize\textwidth][c]{\scriptsize MTL-PoolFormer}\hspace{\qualhspacing}
    \makebox[\qualfigsize\textwidth][c]{\scriptsize MTL-BiSeNetV1}\hspace{\qualhspacing}
    \makebox[\qualfigsize\textwidth][c]{\scriptsize MTL-BiSeNetV2}\hspace{\qualhspacing}
    \makebox[\qualfigsize\textwidth][c]{\scriptsize MTL-SFNet}\hspace{\qualhspacing}
    \makebox[\qualfigsize\textwidth][c]{\scriptsize WeedSense}
    \\[0pt]

    % AMATU row
    \raisebox{3.5ex}{\rotatebox[origin=c]{90}{\textbf{AMATU}}}\hspace{0pt}
    \includegraphics[width=\qualfigsize\textwidth]{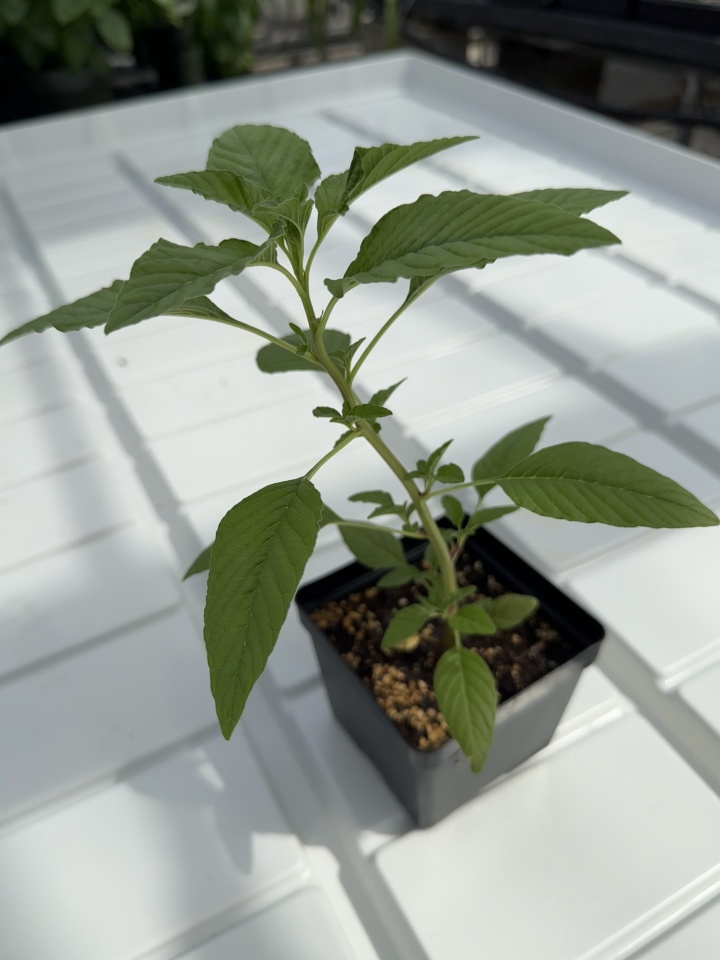}\hspace{\qualhspacing}
    \includegraphics[width=\qualfigsize\textwidth]{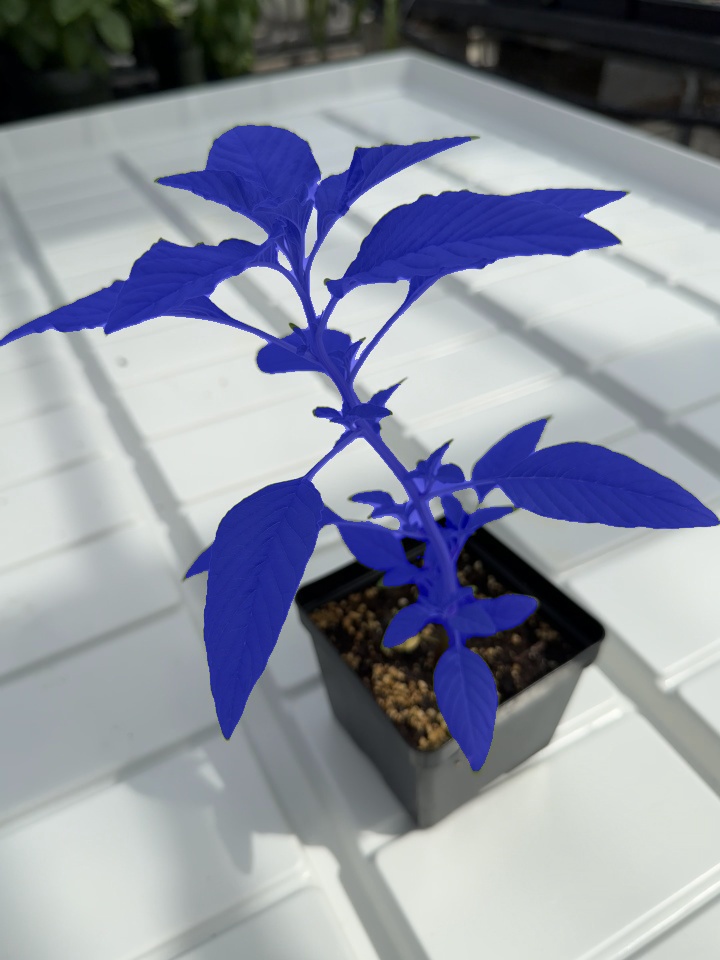}\hspace{\qualhspacing}
    \includegraphics[width=\qualfigsize\textwidth]{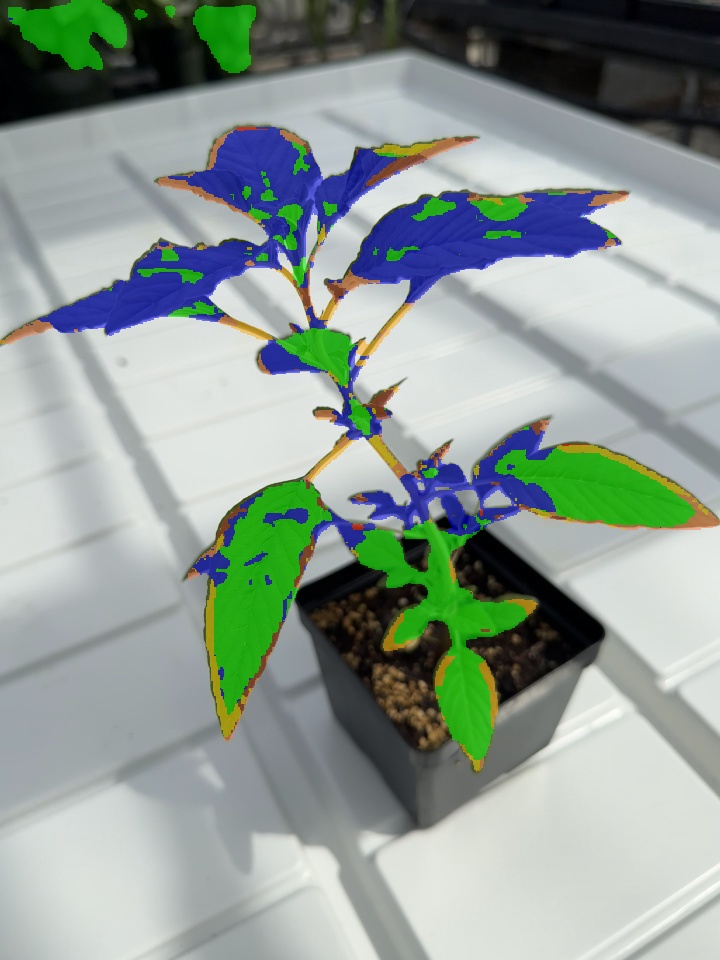}\hspace{\qualhspacing}
    \includegraphics[width=\qualfigsize\textwidth]{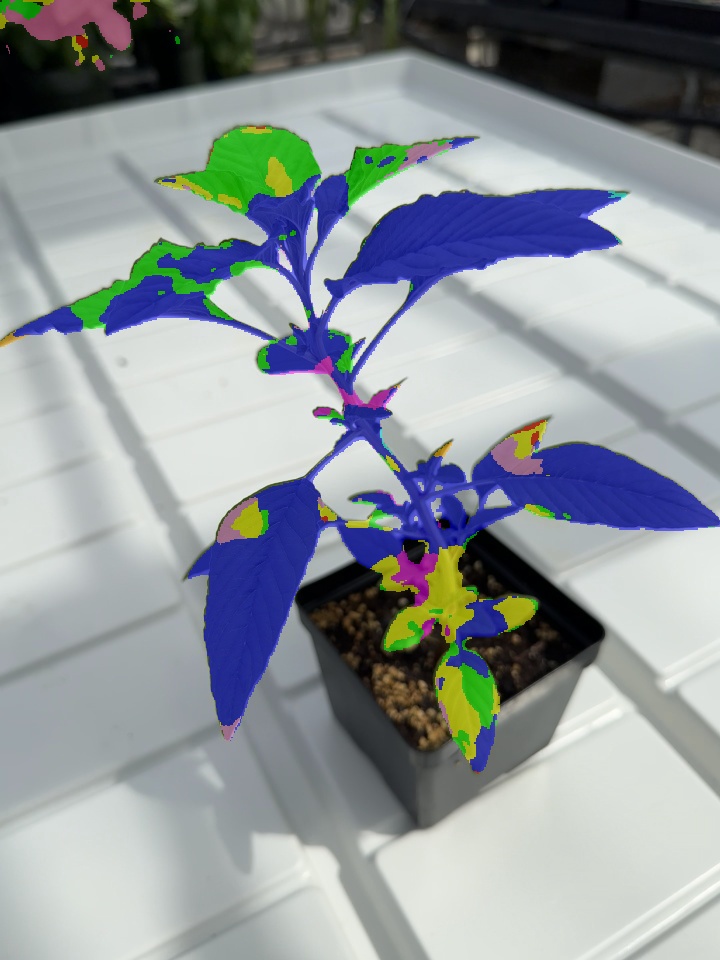}\hspace{\qualhspacing}
    \includegraphics[width=\qualfigsize\textwidth]{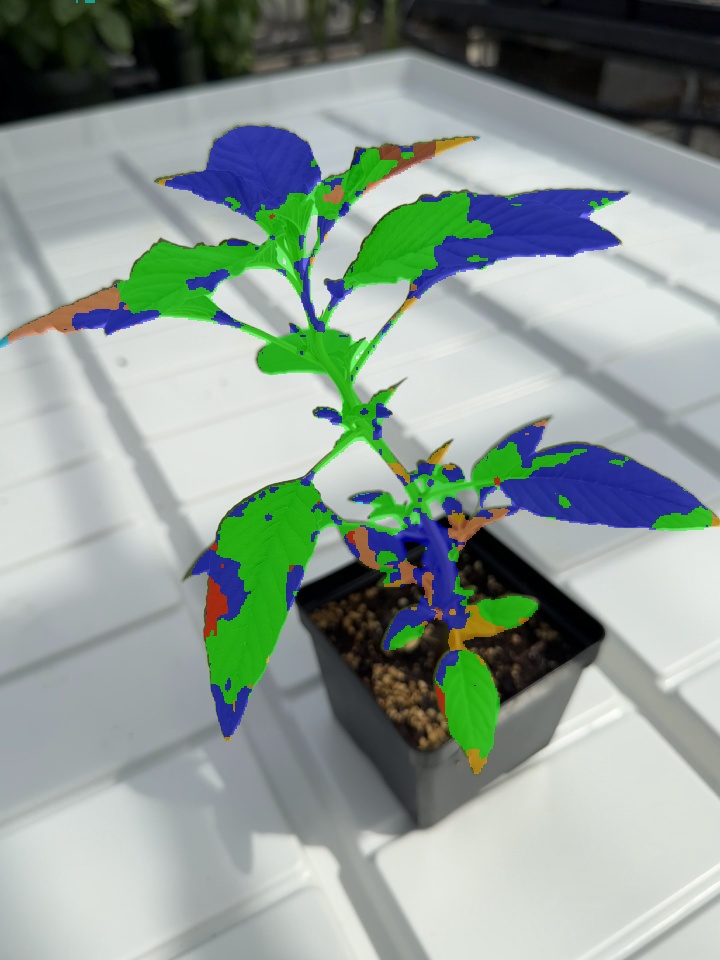}\hspace{\qualhspacing}
    \includegraphics[width=\qualfigsize\textwidth]{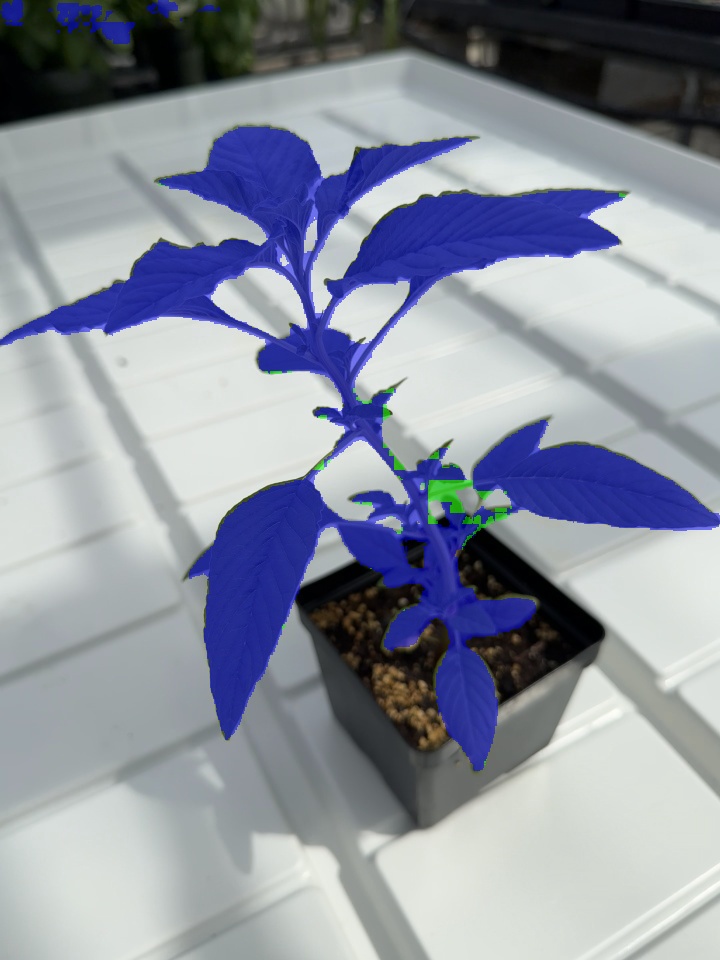}\hspace{\qualhspacing}
    \includegraphics[width=\qualfigsize\textwidth]{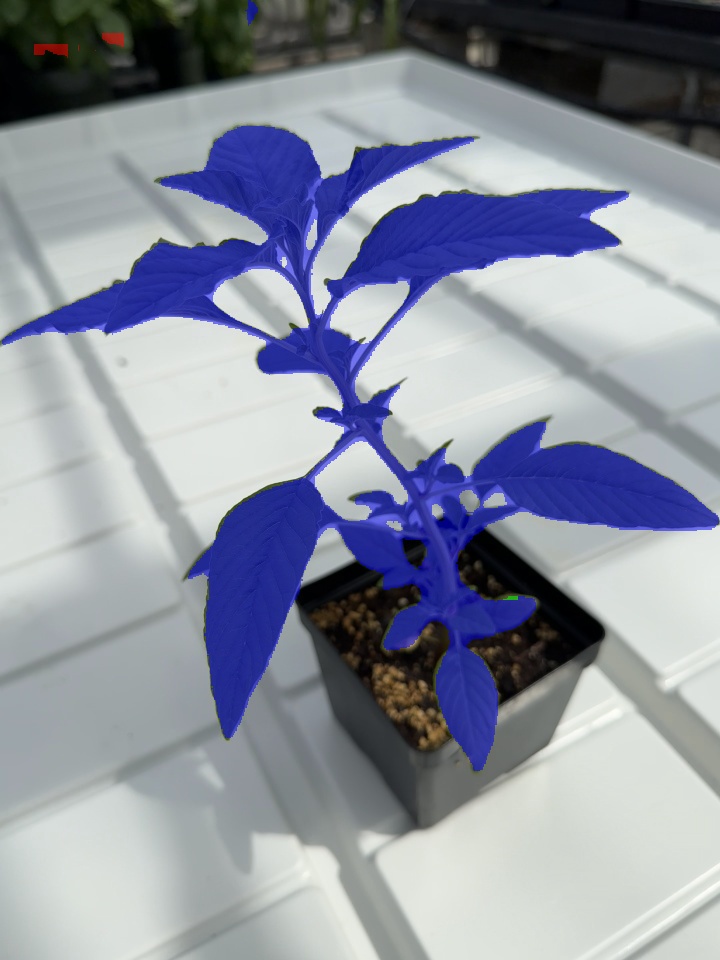}\hspace{\qualhspacing}
    \includegraphics[width=\qualfigsize\textwidth]{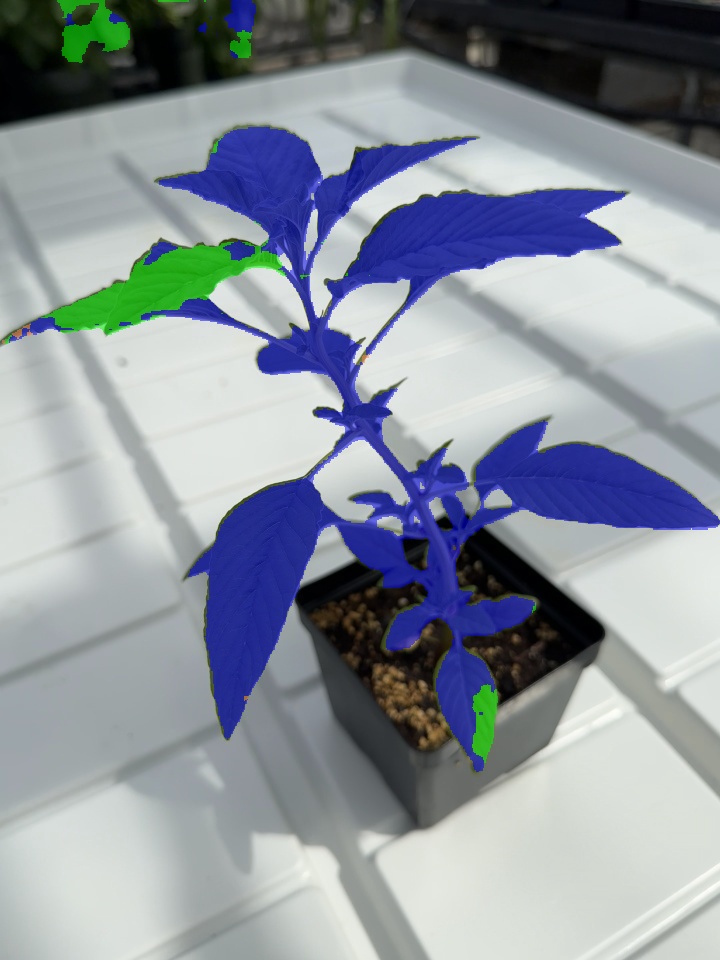}\hspace{\qualhspacing}
    \includegraphics[width=\qualfigsize\textwidth]{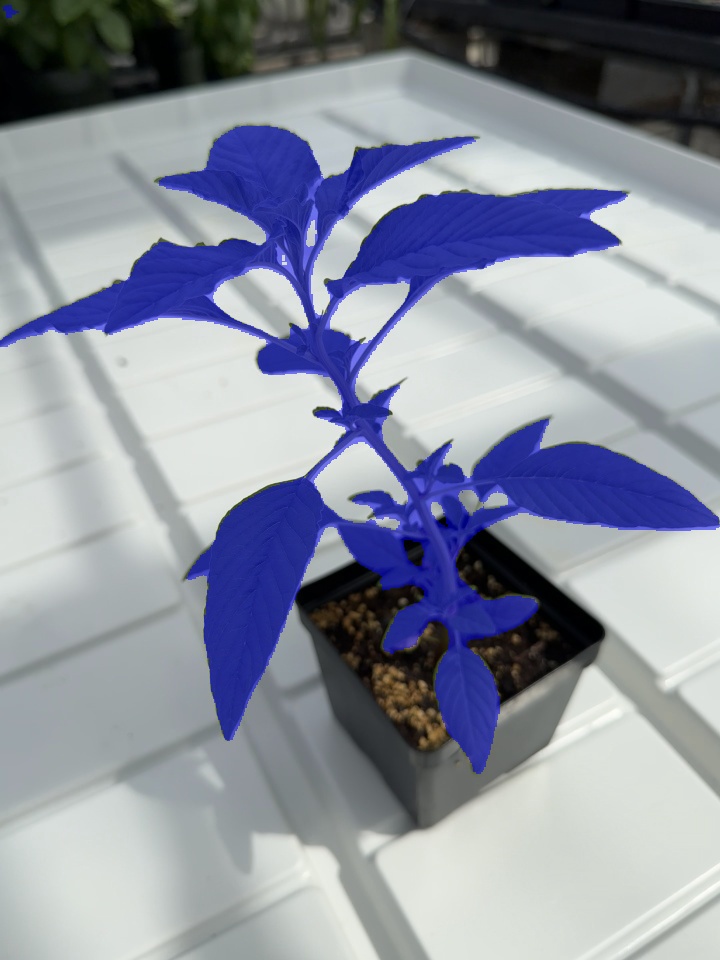}
    \\[0pt]
    
    % CYPES row
    \raisebox{3ex}{\rotatebox[origin=c]{90}{\textbf{CYPES}}}\hspace{0pt}
    \includegraphics[width=\qualfigsize\textwidth]{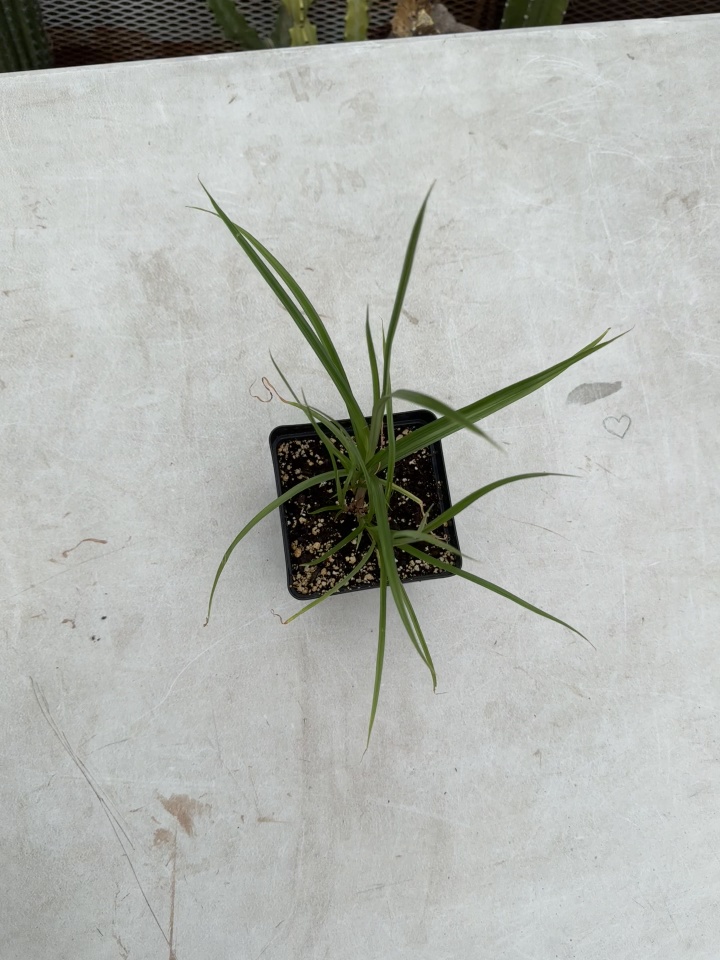}\hspace{\qualhspacing}
    \includegraphics[width=\qualfigsize\textwidth]{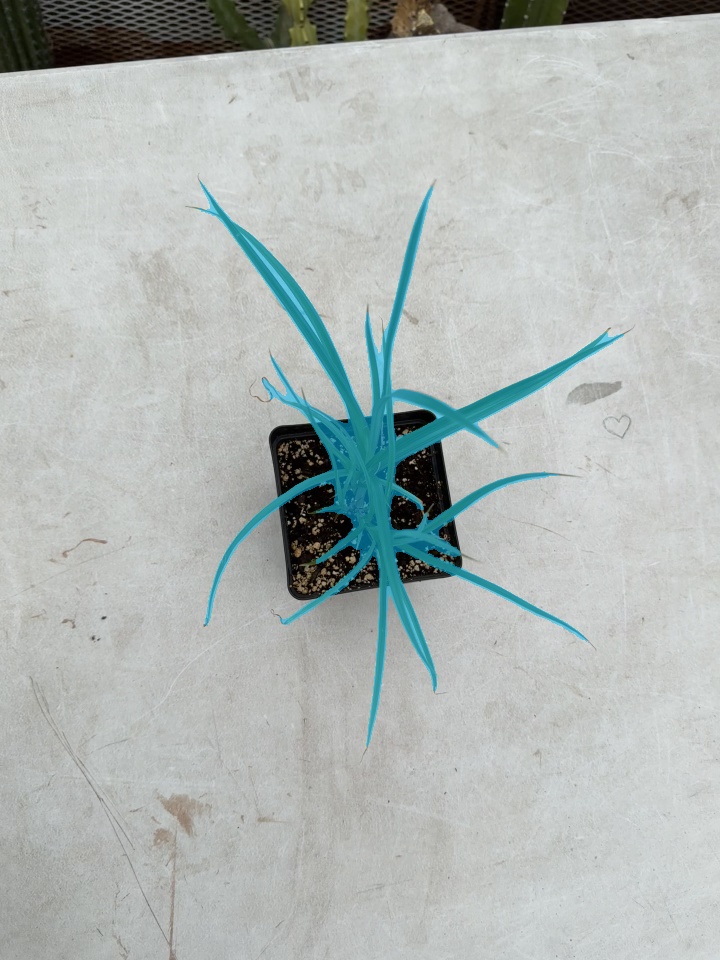}\hspace{\qualhspacing}
    \includegraphics[width=\qualfigsize\textwidth]{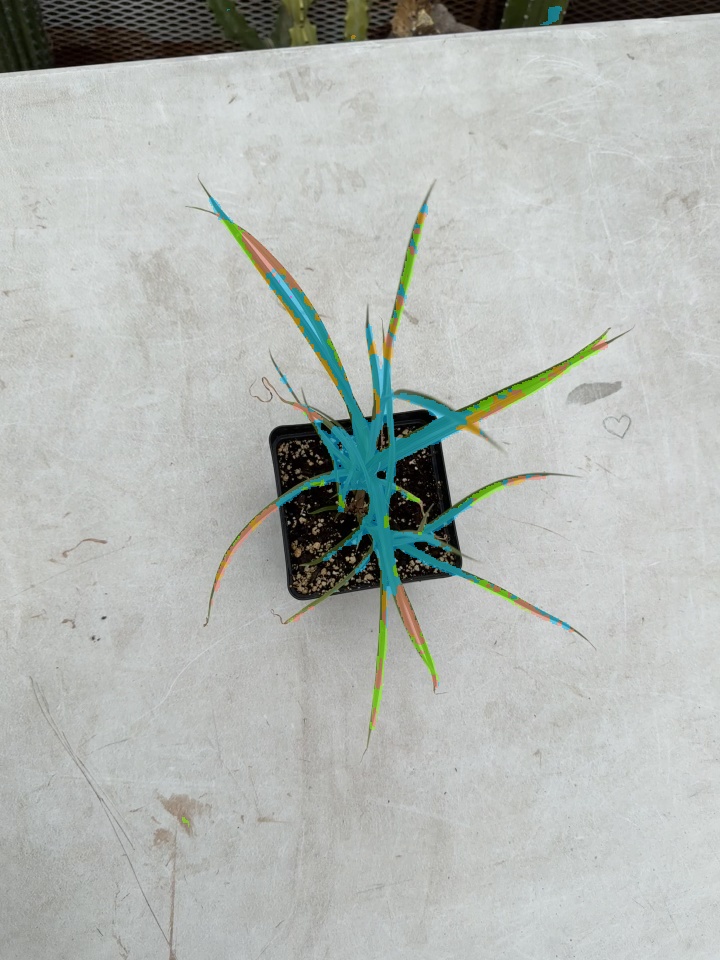}\hspace{\qualhspacing}
    \includegraphics[width=\qualfigsize\textwidth]{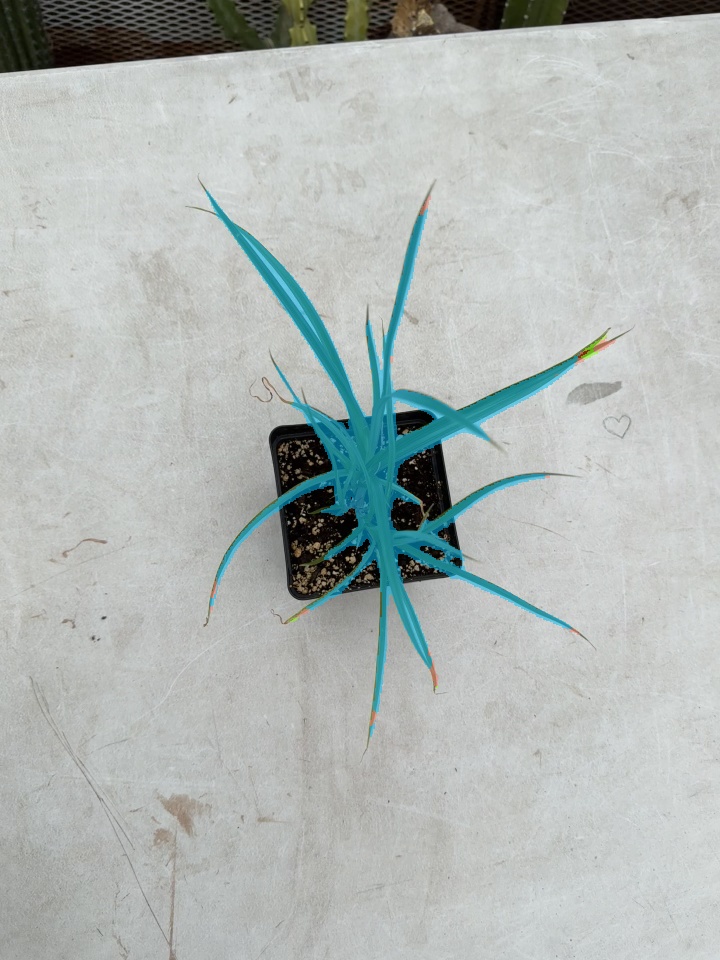}\hspace{\qualhspacing}
    \includegraphics[width=\qualfigsize\textwidth]{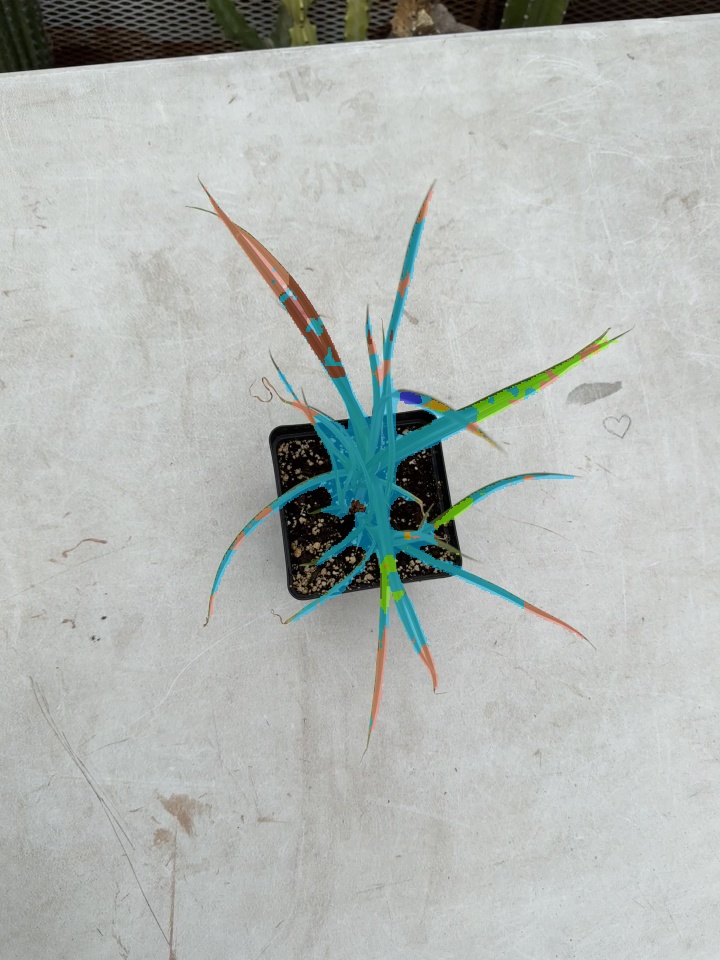}\hspace{\qualhspacing}
    \includegraphics[width=\qualfigsize\textwidth]{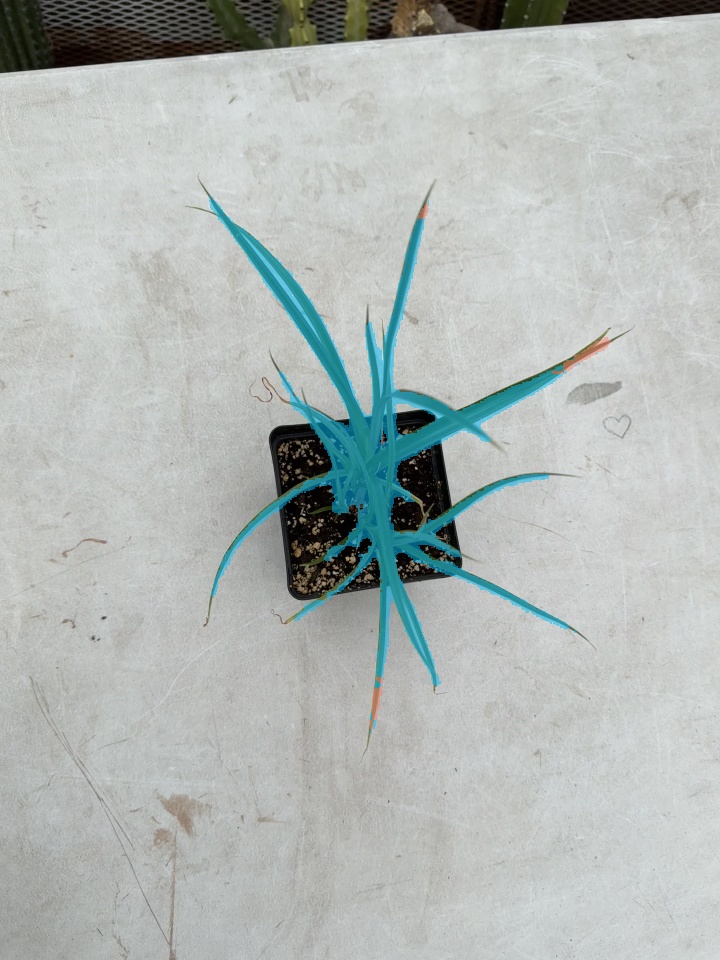}\hspace{\qualhspacing}
    \includegraphics[width=\qualfigsize\textwidth]{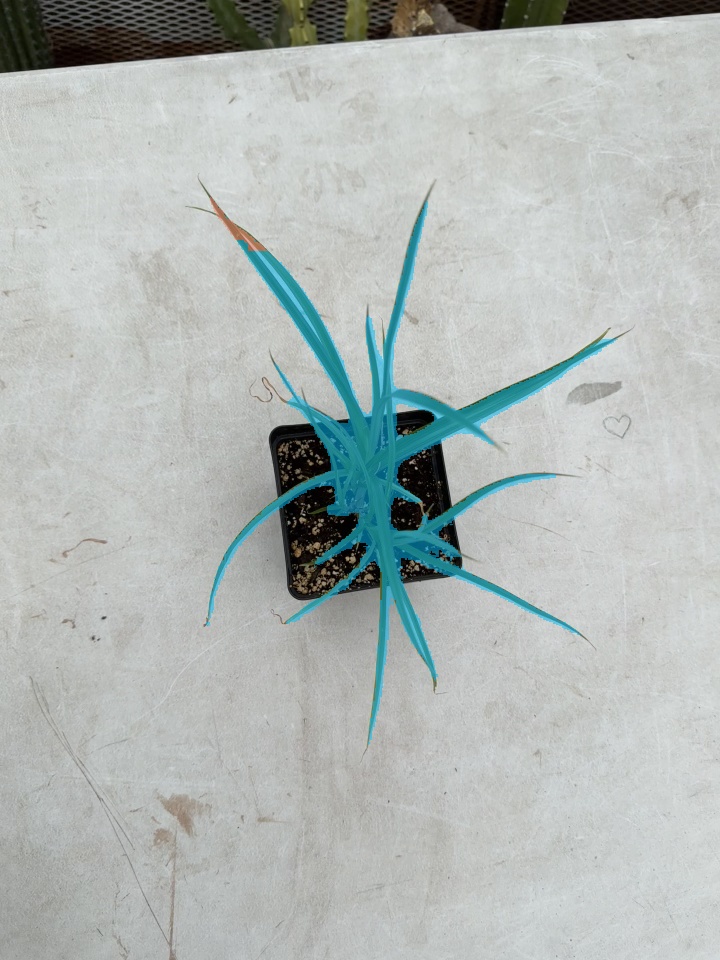}\hspace{\qualhspacing}
    \includegraphics[width=\qualfigsize\textwidth]{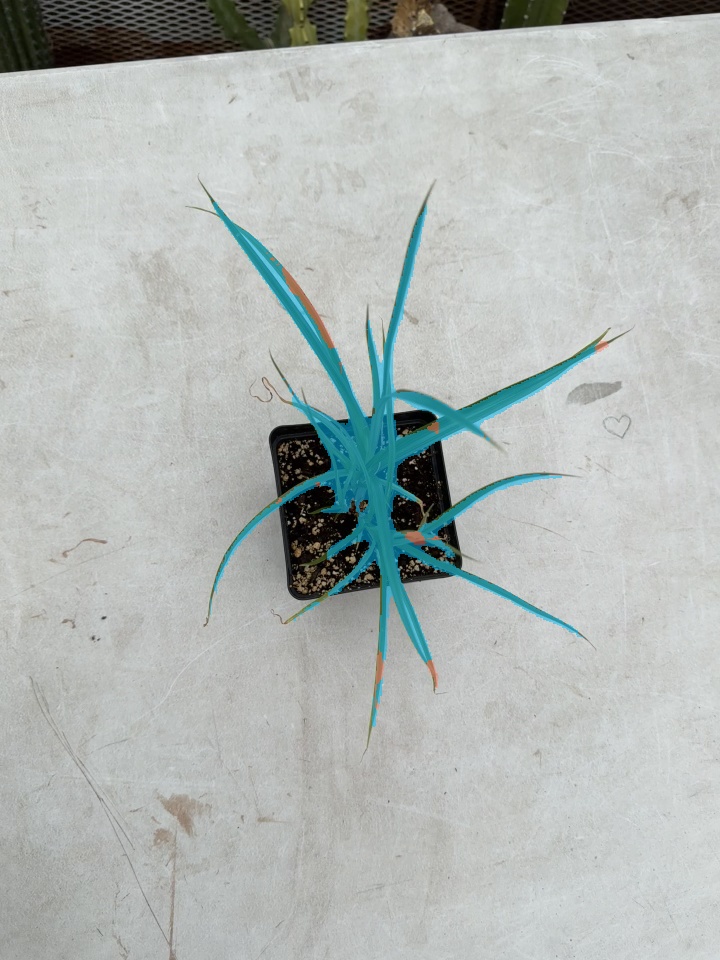}\hspace{\qualhspacing}
    \includegraphics[width=\qualfigsize\textwidth]{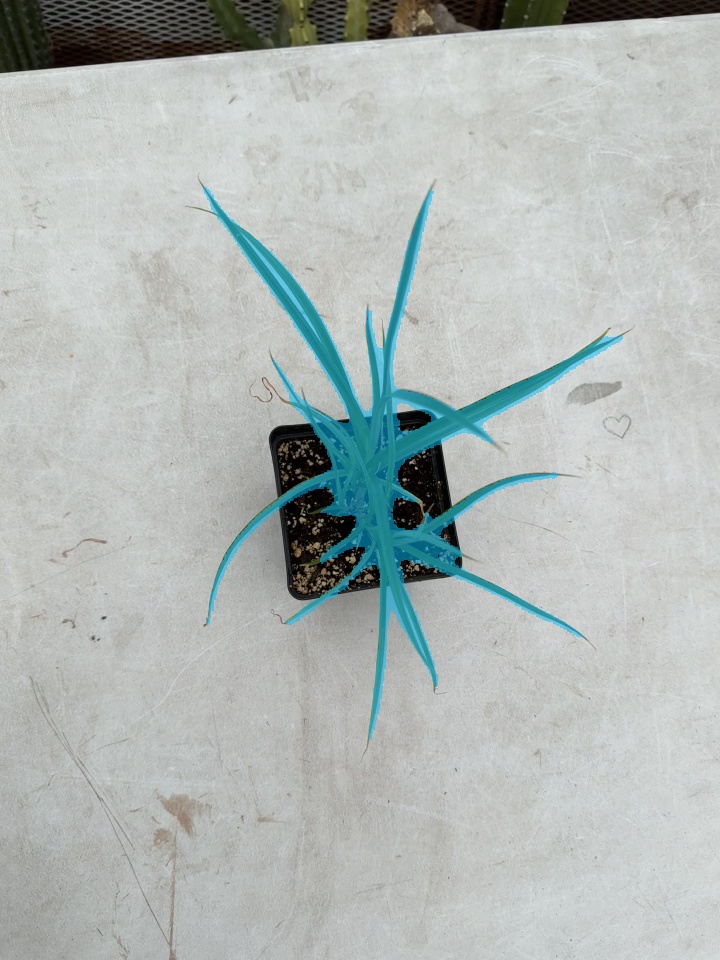}
    \\[0pt]
    
    % Commented out ECHCG row (can be uncommented if needed)
    % ECHCG row
    \raisebox{3ex}{\rotatebox[origin=c]{90}{\textbf{ECHCG}}}\hspace{0pt}
    \includegraphics[width=\qualfigsize\textwidth]{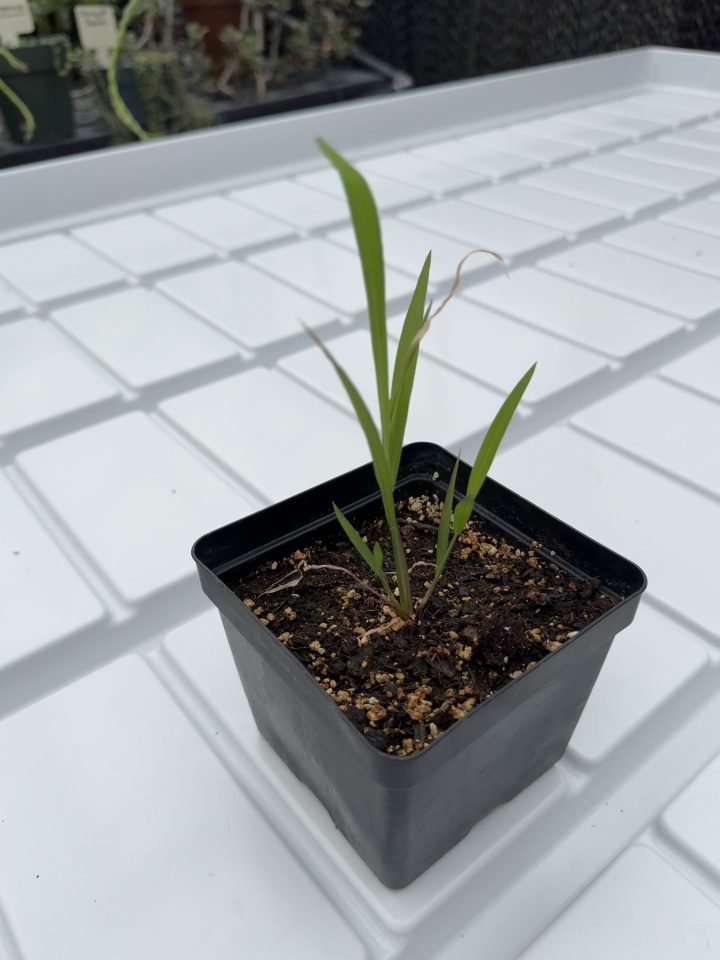}\hspace{\qualhspacing}
    \includegraphics[width=\qualfigsize\textwidth]{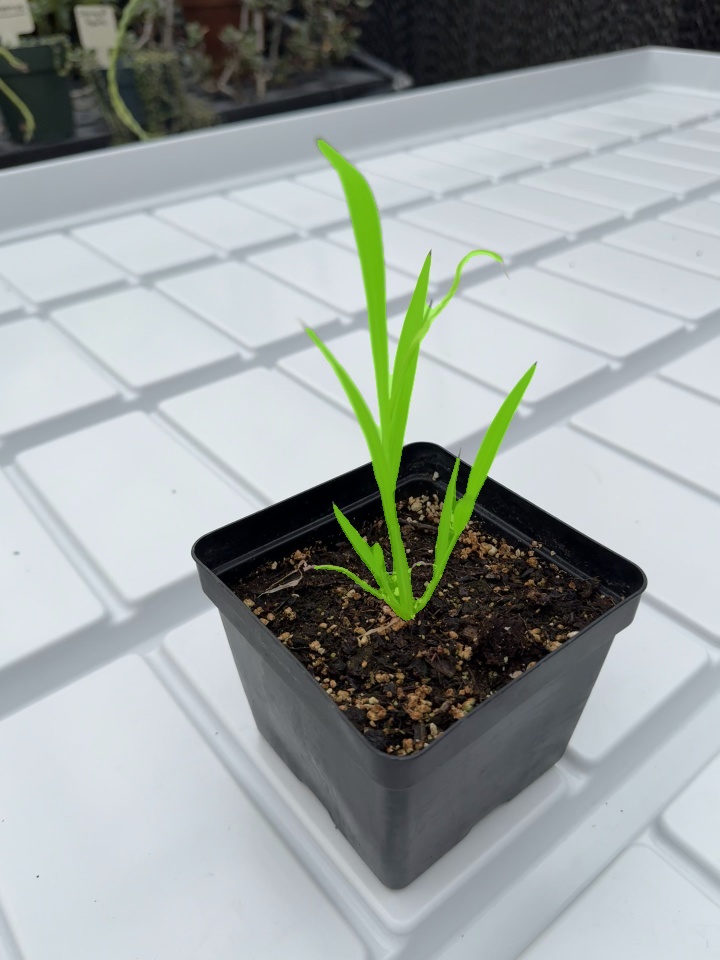}\hspace{\qualhspacing}
    \includegraphics[width=\qualfigsize\textwidth]{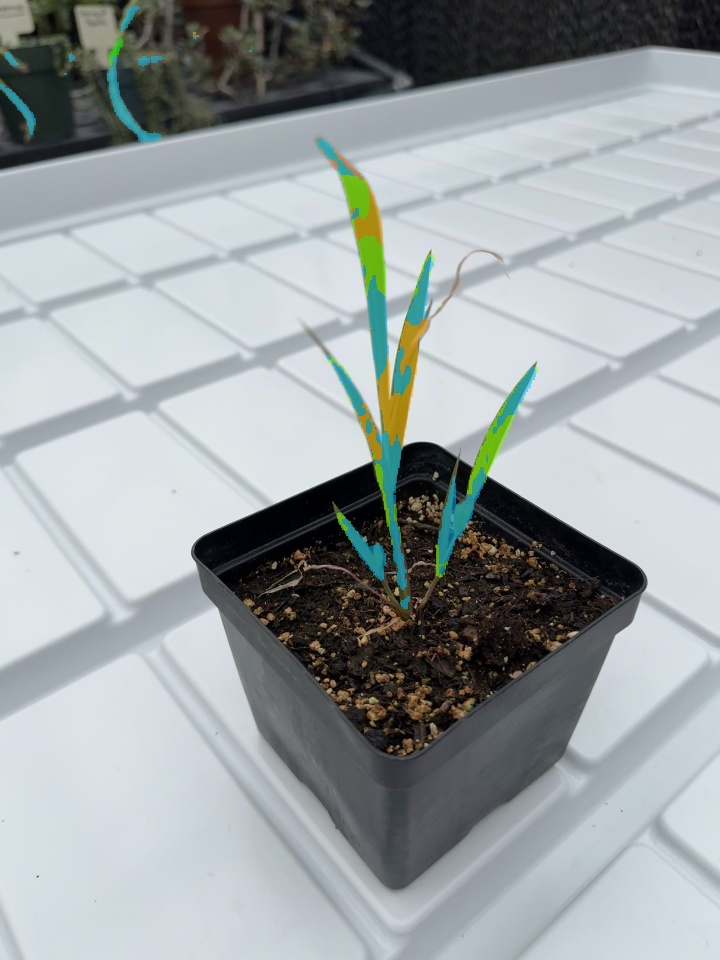}\hspace{\qualhspacing}
    \includegraphics[width=\qualfigsize\textwidth]{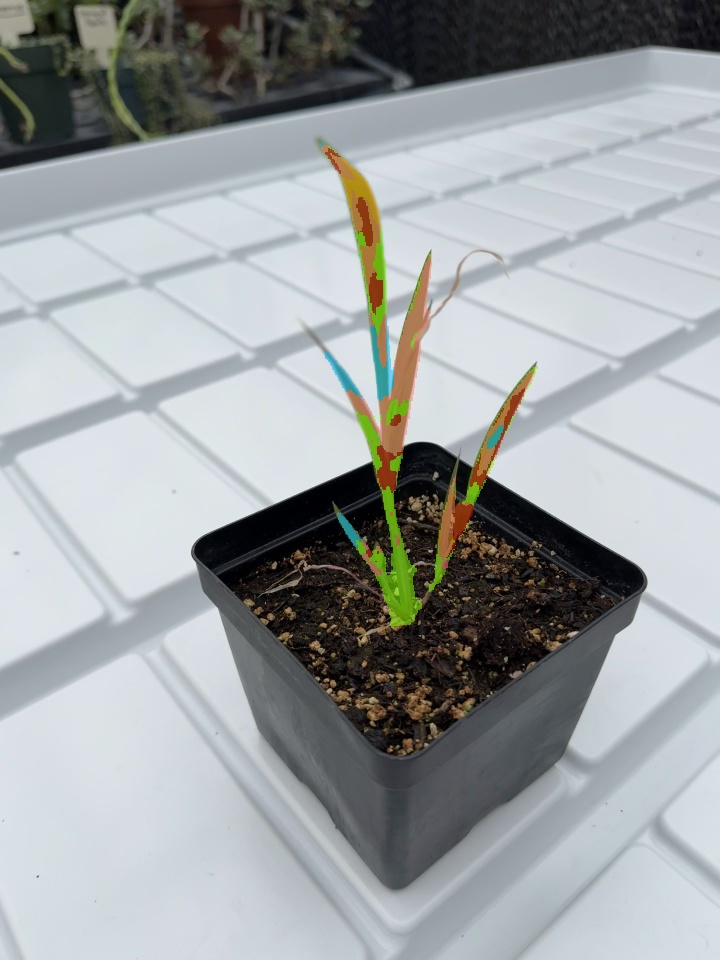}\hspace{\qualhspacing}
    \includegraphics[width=\qualfigsize\textwidth]{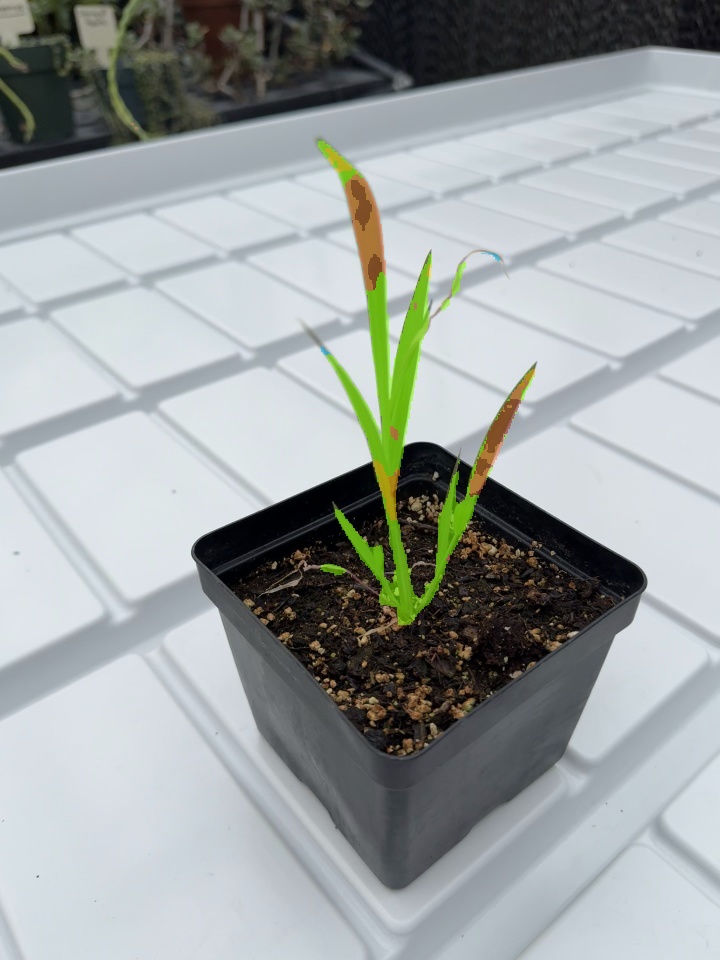}\hspace{\qualhspacing}
    \includegraphics[width=\qualfigsize\textwidth]{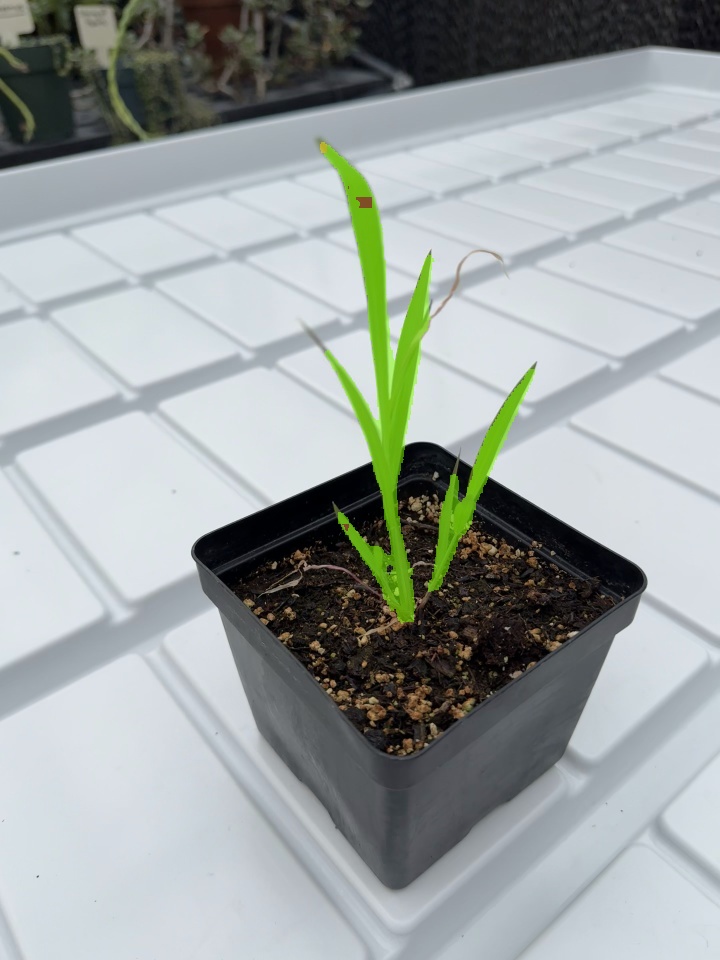}\hspace{\qualhspacing}
    \includegraphics[width=\qualfigsize\textwidth]{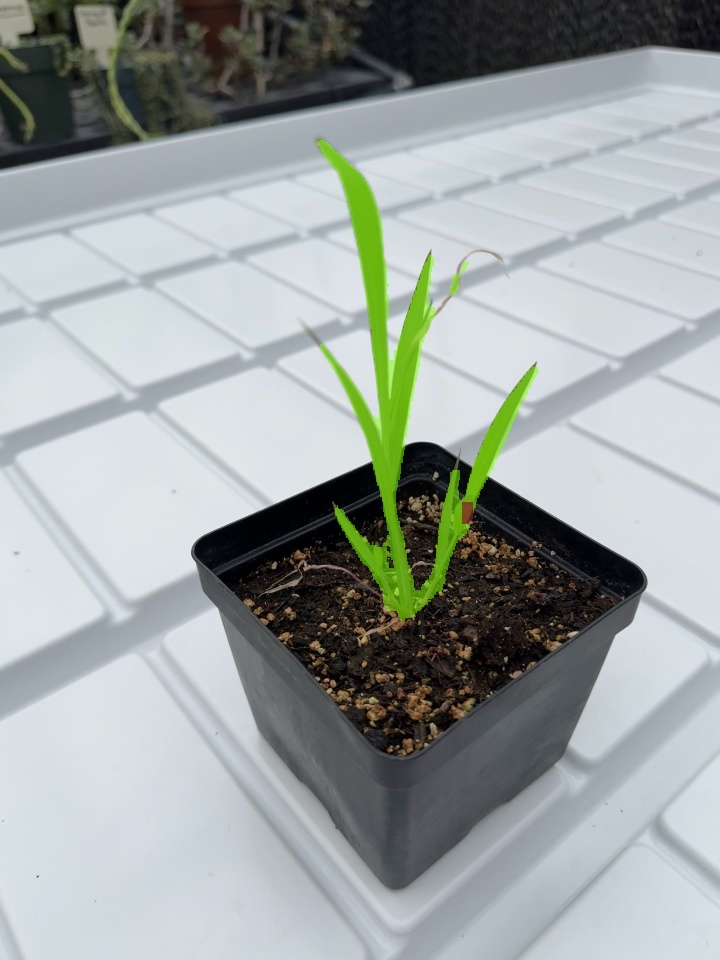}\hspace{\qualhspacing}
    \includegraphics[width=\qualfigsize\textwidth]{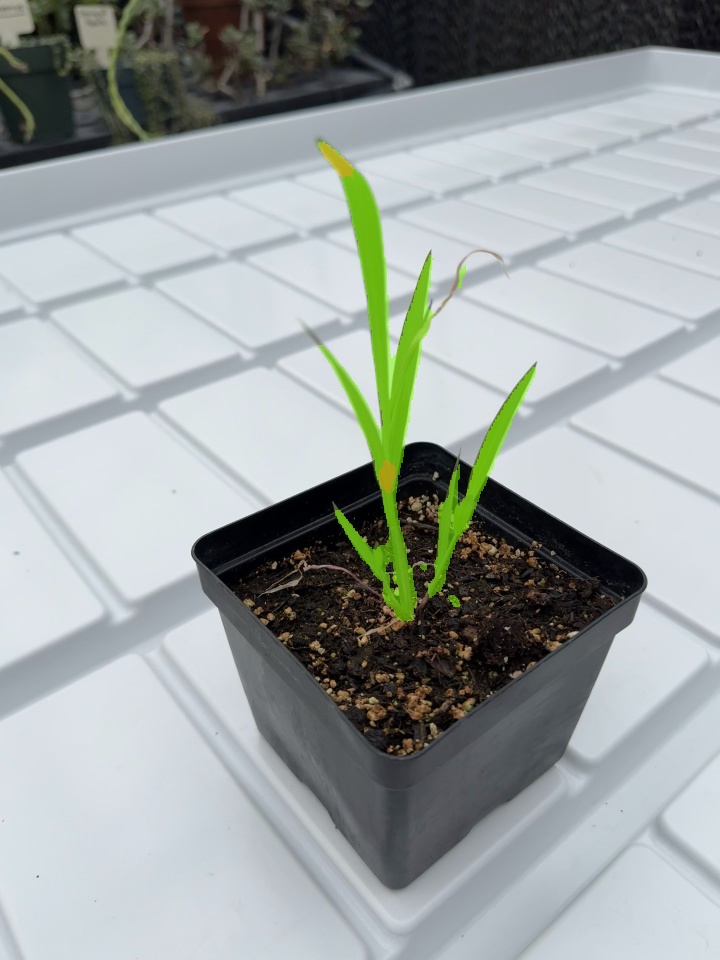}\hspace{\qualhspacing}
    \includegraphics[width=\qualfigsize\textwidth]{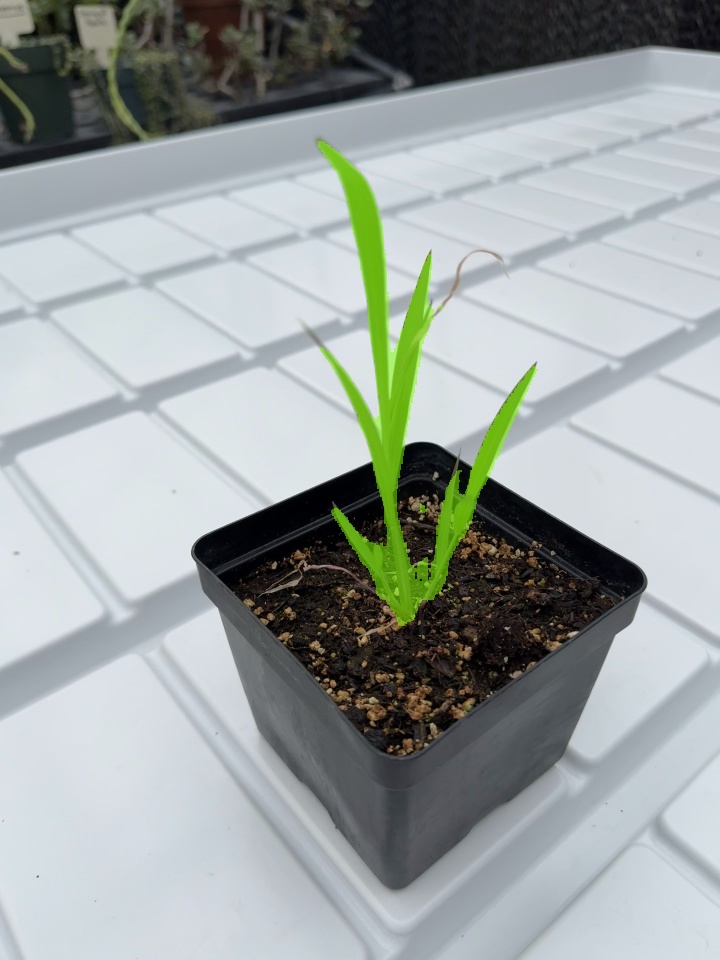}
    
    \caption{Qualitative comparison of different models, showing the input images, ground truth masks, and the segmentation outputs for AMATU, CYPES and ECHCG. Best viewed on screen.}
    \label{fig:qualitative-comparison}
    \vspace{-15pt}
\end{figure*}

\noindent \textbf{Growth Stage Classification Performance.} As shown in ~\Cref{tab:performance_comparison}, WeedSense achieves 99.99\% accuracy and F1 score on growth stage classification. The top-performing models achieve near-perfect results, with WeedSense ranking among the best performers alongside MTL-BiSeNetV2 and MTL-BiSeNetV1. Performance differences among top models are negligible for practical applications.
The consistently high classification performance across most architectures indicates that growth stage classification is more tractable than segmentation or height estimation, likely due to distinctive visual patterns associated with different growth stages.

\noindent \textbf{Real-Time Performance Analysis.} As shown in ~\Cref{tab:comp_efficiency}, WeedSense achieves 160 FPS with 30.50M parameters and 16.73 GFLOPs, providing a balance between performance and efficiency. While MTL-BiSeNetV1 offers the highest frame rate at 249 FPS and MTL-SegFormer requires the fewest resources at 7.94 GFLOPs, WeedSense delivers competitive overall efficiency-performance trade-offs.
WeedSense runs 1.7× faster than MTL-UNet while requiring 13.94× fewer computational resources and achieving 38.27 percentage points better segmentation mIoU. Compared to MTL-SFNet, our approach provides 1.84× computational efficiency improvement while maintaining 3.51 percentage points better segmentation performance.

% \vspace{-5pt}
\subsection{Ablation Studies}%
\label{sec:ablation}

To validate our architectural design choices and determine the optimal configuration for WeedSense, we conduct extensive ablation experiments. 
All experiments follow the training protocol described in \cref{sec:implementation_details}.

\noindent \textbf{Kernel Configuration.} We first analyze the effect of kernel size configuration for UIB blocks, denoted as S[start]-M[mid]-E[end]. \Cref{tab:ablation_table} shows the performance results across different configurations. The first thing to observe is that the S5-M3-E5 configuration achieves the highest segmentation mIoU at 89.53\% but requires slightly more computational resources. The S0-M3-E0 configuration maintains competitive performance while offering the lowest computational cost. For height estimation, the S0-M3-E5 configuration achieves the lowest MAE at 1.64cm, while S1-M3-E0 provides competitive performance at 1.65cm. Interestingly, all configurations achieve near-perfect growth stage classification accuracy. From these results, we select the S0-M3-E0 configuration as it provides the optimal efficiency-performance trade-off.

\noindent \textbf{Squeeze-and-Excitation Module Effects.} Building upon the S0-M3-E0 configuration, we analyze the impact of squeeze-and-excitation (SE) modules. As shown in \Cref{tab:ablation_table}, incorporating SE modules increase segmentation mIoU by 0.94 percentage points and reduces height estimation error by 0.07 cm. This performance gain comes at the cost of only 1.08M additional parameters with no change in computational complexity. From these results, we can conclude that adaptive channel recalibration improves feature quality with minimal computational overhead.

\noindent \textbf{Channel Capacity.}
We now analyze the effect of channel capacity in the aggregation layer and SE modules on model performance. \Cref{tab:ablation_table} shows performance across different configurations. The first thing to observe is that adding SE modules universally improves performance regardless of channel dimensions. When comparing configurations, the 256-channel with SE achieves better height estimation but similar segmentation performance compared to the 128-channel variant, yet requires 45.3\% more computation. For segmentation, performance increases as channel capacity increases up to 128-channels, then plateaus with minimal gains at 256-channels. For height estimation, performance continues to improve with wider channels. The 64-channel variants show substantial performance degradation despite their efficiency advantages. From these results, we select the 128-channel with SE configuration as it offers the best balance between accuracy and efficiency for deployment.

\begin{table}[tpb]
\vspace{-5pt}
% \resizebox{0.48\textwidth}{!}{
\centering
\scriptsize
\setlength{\tabcolsep}{3pt}
\begin{tabular}{l|ccc|cc}
\toprule
% \textbf{Config.}& \textbf{Segmentation} & \textbf{Height} & \textbf{Week} & \multicolumn{2}{c}{\textbf{Comp. Efficiency}} \\
% \midrule
Config. & \makecell{Seg.$\uparrow$\\ mIoU (\%)} & \makecell{Height$\downarrow$\\ MAE (cm)} & \makecell{Week$\uparrow$ \\ Acc (\%)} & \makecell{ \\ Params (M)$\downarrow$} & \makecell{ \\ GFLOPs$\downarrow$} \\
\midrule

\multicolumn{6}{l}{\textit{UIB Kernel Configuration}} \\
\midrule
\rowcolor{gray!20}
S0-M3-E0 & 88.84 & 1.74 & 100.00 & 29.42 & 16.73 \\
S1-M3-E0 & 88.81 & 1.65 & 100.00 & 29.43 & 16.74 \\
S0-M3-E1 & 89.43 & 1.67 & 100.00 & 29.43 & 16.74 \\
S1-M3-E1 & 89.49 & 1.75 & 99.98 & 29.43 & 16.74 \\
S5-M3-E0 & 89.41 & 1.67 & 99.99 & 29.44 & 16.75 \\
S0-M3-E5 & 89.17 & 1.64 & 99.91 & 29.44 & 16.75 \\
S5-M3-E5 & 89.53 & 1.70 & 99.96 & 29.46 & 16.77 \\
\midrule
\multicolumn{6}{l}{\textit{Squeeze-and-excitation Module}} \\
\midrule
No SE & 88.84 & 1.74 & 100.00 & 29.42 & 16.73 \\
\rowcolor{gray!20}
With SE & 89.78 & 1.67 & 99.99 & 30.50 & 16.73 \\
\midrule
\multicolumn{6}{l}{\textit{Channel Capacity Configuration}} \\
\midrule
C64-NoSE & 86.50 & 2.28 & 99.87 & 28.45 & 13.70 \\
C64-SE & 87.31 & 2.00 & 99.99 & 29.53 & 13.70 \\
C128-NoSE & 88.84 & 1.74 & 100.00 & 29.42 & 16.73 \\
\rowcolor{gray!20}
C128-SE & 89.78 & 1.67 & 99.99 & 30.50 & 16.73 \\
C256-NoSE & 89.37 & 1.52 & 100.00 & 32.13 & 24.31 \\
C256-SE & 89.70 & 1.52 & 100.00 & 33.21 & 24.31 \\
\midrule
\multicolumn{6}{l}{\textit{Auxiliary Supervision Configuration}} \\
\midrule
No Aux & 87.48 & 2.05 & 98.70 & 9.62 & 16.73 \\
\rowcolor{gray!20}
Aux S1-4 & 89.78 & 1.67 & 99.99 & 30.50 & 16.73 \\
\midrule
\multicolumn{6}{l}{\textit{Model Size Configuration}} \\
\midrule
Small & 77.17 & 2.66 & 99.69 & 21.50 & 3.56 \\
\rowcolor{gray!20}
Medium & 89.78 & 1.67 & 99.99 & 30.50 & 16.73 \\
Large & 91.41 & 1.41 & 99.78 & 45.36 & 34.55 \\
\midrule
\multicolumn{6}{l}{\textit{Task Configuration}} \\
\midrule
Single Task & 93.67 & 1.53 & 99.55 & - & - \\
\rowcolor{gray!20}
Multi Task & 89.78 & 1.67 & 99.99 & - & - \\
\bottomrule
\end{tabular}
% }
\vspace{-5pt}
\caption{Comprehensive ablation analysis of WeedSense design components. Results demonstrate the progressive refinement from baseline S0-M3-E0 kernel configuration through the addition of squeeze-and-excitation modules, optimal channel capacity selection, auxiliary supervision, and model size scaling. The final comparison shows trade-offs between single-task and multi-task learning approaches.}
\label{tab:ablation_table}
\vspace{-20pt}
\end{table}

\noindent \textbf{Auxiliary Supervision.} ~\Cref{tab:ablation_table} shows the results with and without auxiliary heads. Removing auxiliary heads reduces segmentation mIoU by 2.30 percentage points and increases height estimation error by 22.75\%. Growth stage classification accuracy also drops from 99.99\% to 98.70\%. 
As described in \cref{sec:dual-path-encoder}, these auxiliary components are training-only and incur zero computational overhead during inference.

\noindent \textbf{Model Size Scaling.} We implement three model variants that systematically scale key architectural components to analyze capacity-performance trade-offs. \Cref{tab:model_scaling_config} details the scaling strategy across both encoder branches and decoder components.
As shown in \Cref{tab:ablation_table}, performance increases predictably with model size. The Large variant improves segmentation mIoU by 1.63 percentage points and reduces height MAE by 0.26 cm compared to Medium. However, these gains come at the cost of significantly increased computational requirements. In contrast, the Small variant offers substantially lower computational cost but with substantial accuracy degradation compared to Medium. From these results, we conclude that the Medium variant provides the optimal compromise for practical deployment.

\noindent \textbf{Single vs. Multitask.} We compare our multitask approach against dedicated single-task models using the same Medium-variant architecture. \Cref{tab:ablation_table} shows that while single-task models achieve 3.89\% better segmentation mIoU and 0.14 cm lower height MAE, our multitask model provides substantial computational advantages. Individual single-task models require 45.14M total parameters (segmentation: 30.50M, week: 7.32M, height: 7.32M) compared to our multitask model's 30.50M parameters, representing a 32.4\% parameter reduction. For inference efficiency, sequential execution of single-task models achieves only 52.3 FPS, while parallel execution reaches 157.84 FPS but requires significantly more resources. Our unified approach achieves 160 FPS with a single model, outperforming even parallel single-task execution while using substantially fewer resources, demonstrating an optimal balance between performance and efficiency for practical deployment.

\begin{table}[tpb]
\vspace{-5pt}
\resizebox{0.48\textwidth}{!}{
\centering
\scriptsize
\setlength{\tabcolsep}{4pt}
\begin{tabular}{lccc}
\toprule
Component & Small & Medium & Large \\
\midrule
Detail Branch Channels & 32-32-64 & 64-64-128 & 96-96-192 \\
Semantic Branch Channels & 8-16-32-64 & 16-32-64-128 & 24-48-96-192 \\
UIB Blocks per Stage & 1-1-2 & 2-2-4 & 3-3-6 \\
Expansion Ratio & 4 & 6 & 6 \\
Transformer Embed Dim & 256 (4 heads) & 512 (8 heads) & 768 (12 heads) \\
Task Head Dimensions & 512→256 & 1024→512 & 1536→768 \\
\bottomrule
\end{tabular}
}
\vspace{-5pt}
\caption{Model size variant configurations. Branch channels correspond to stages S1-S2-S3 (Detail) and stem-S3-S4-S5 (Semantic). UIB blocks are distributed across S3-S4-S5 stages.}
\label{tab:model_scaling_config}
\vspace{-15pt}
\end{table}
\vspace{-5pt}
\section{Conclusion}
% \vspace{-5pt}

In this study, we proposed WeedSense, a novel multi-task learning architecture for comprehensive weed analysis that simultaneously performs semantic segmentation, height estimation, and growth stage classification using RGB imagery. We also introduced a novel dataset capturing 16 weed species over an 11-week growth cycle with pixel-level annotations, height measurements, and weekly growth stage labels to evaluate the model's performance across diverse growth patterns and species characteristics. Experimental evaluation demonstrated that WeedSense outperforms state-of-the-art models across key performance metrics including segmentation accuracy, height estimation precision, and growth stage classification using weekly intervals. 
Our unified multitask approach achieves computational efficiency gains, requiring 32.4\% fewer parameters than separate single-task models while maintaining 160 FPS inference speed compared to 52.3 FPS for sequential single-task execution.
Future research could explore the integration of WeedSense with morphology-based BBCH growth stage classification, field-based data collection to validate performance in real agricultural environments, and crop-weed discrimination tasks to enhance its adaptability in diverse agricultural environments while investigating its robustness across different lighting conditions and weather scenarios.

{
    \small
    \bibliographystyle{ieeenat_fullname}
    \bibliography{main}
}

\end{document}